\documentclass{article}
    \PassOptionsToPackage{numbers, compress}{natbib}

    \usepackage[preprint]{neurips_2022}

\usepackage[utf8]{inputenc} 
\usepackage[T1]{fontenc}    
\usepackage[colorlinks,linkcolor=blue,urlcolor=blue,citecolor=blue]{hyperref}
\usepackage{url}            
\usepackage{booktabs}       
\usepackage{amsfonts}       
\usepackage{nicefrac}       
\usepackage{microtype}      

\usepackage{xspace}
\usepackage{xcolor}
\usepackage{stmaryrd}
\usepackage{amsthm}
\usepackage{amsmath}
\usepackage{amssymb}
\usepackage{mathtools}
\usepackage[pdftex]{graphicx}
\graphicspath{{paper_figures/}}
\usepackage{wrapfig}
\usepackage{subcaption}
\usepackage{multirow}

\usepackage{algorithm}
\usepackage{algorithmic}
\usepackage{arydshln}
\usepackage{threeparttable}

\newcommand{\ie}{\textit{i.e.}}
\newcommand{\eg}{\textit{e.g.}}

\newcommand{\hidecomments}[1]{}

\usepackage{amsmath,amsfonts,bm}

\def\eqref#1{equation~\ref{#1}}

\def\1{\bm{1}}

\DeclareMathAlphabet{\mathsfit}{\encodingdefault}{\sfdefault}{m}{sl}
\SetMathAlphabet{\mathsfit}{bold}{\encodingdefault}{\sfdefault}{bx}{n}

\newcommand{\sigmoid}{\sigma}

\newcommand{\xhdr}[1]{{\noindent\bfseries #1}.}

\newcommand{\cut}[1]{}

\newtheorem{proposition}{Proposition}

\newcommand{\repourl}{\url{https://github.com/LinXueyuanStdio/BoolE}}
\renewcommand{\repourl}{\url{https://anonymous.4open.science/r/FLEX_NIPS}}

\title{FLEX: \textsl{F}eature-\textsl{L}ogic \textsl{E}mbedding Framework\\
    for Comple\textsl{X} Knowledge Graph Reasoning}

\author{
  Xueyuan Lin, Haihong E\thanks{Corresponding Author}, Gengxian Zhou, Tianyi Hu\\
  \textbf{Ningyuan Li, Mingzhi Sun, Haoran Luo}\\
  Department of Computer Science, \\
  Beijing University of Posts and Telecommunications, Beijing, China\\
  \texttt{\{linxy59, ehaihong, 2018213090, hutianyi\}@bupt.edu.cn} \\
  \texttt{\{jason.ningyuan.li, sunmingzhi, luohaoran\}@bupt.edu.cn} \\
}

\begin{document}

\maketitle

\begin{abstract}
Current best performing models for knowledge graph reasoning (KGR) introduce geometry objects or probabilistic distributions to embed entities and first-order logical (FOL) queries into low-dimensional vector spaces.
However, they have limited logical reasoning ability, because not all logical operations are closed.
And it is difficult to generalize to various features, unable to utilize heterogeneous information.
To address these challenges, we propose a closed embedding-based framework, \textsl{F}eature-\textsl{L}ogic \textsl{E}mbedding framework, which not only supports various feature spaces, but also has all closed logical operators, including conjunction, disjunction, negation, etc.
Specifically, we introduce vector logic, which naturally models all FOL operations.
The experimental results demonstrate that FLEX is a promising framework which has stable performance, extensibility, flexibility, and strong logical reasoning capability.
\end{abstract}

\section{Introduction}\label{sec:introduction}

\begin{figure*}
    \centering
    \vspace{-3mm}
    \includegraphics[width=\textwidth]{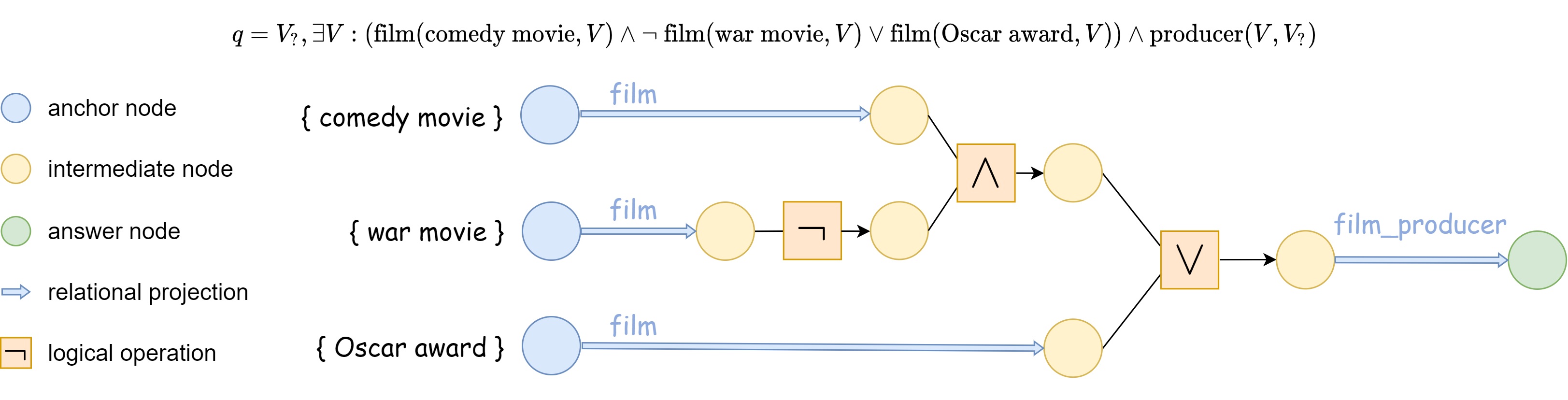}
    \caption{A multi-hop logical query and its computation graph.}
    \vspace{-5mm}
    \label{fig:query}
\end{figure*}

One of the fundamental tasks in knowledge graphs (KGs) is knowledge graph reasoning (KGR), which aims to infer new facts by performing complex multi-hop logical reasoning over the given KG.
Since the KGs are usually large and incomplete, the KGR task is challenging.
In general, multi-hop logical reasoning answers a first-order logic (FOL) query involving logical and relational operators.
Recently, the most popular models for logical reasoning are embedding-based.
These methods represent the query as low-dimensional embedding, which are used to rank against all candidate entities to find the answers.
Figure~\ref{fig:query} shows a multi-hop query example and its computation graph.

Current common embedding-based methods include:
(1) distribution-based model such as: BetaE~\cite{BetaE} and PERM~\cite{PERM},
(2) geometry-based model such as Q2B~\cite{Query2box}, ConE~\cite{ConE}, HypE~\cite{HypE}.
These methods can be characterized as a center-size framework, where the center parameter represents semantic position and the size parameter represents boundary.
For example, in PERM, center is the mean value and size is covariance value of the Gaussian distribution.
In Q2B, center is the center of box, while size is the offset of the box.
In ConE, center is the axis and size is the aperture.

However, we have identified two issues with the center-size framework.
First, \textbf{not all logical operations are closed under this framework}.
For example, it is not possible to use the Beta distribution to represent the difference operation, because the difference of two Beta distributions is not a Beta distribution.
Additionally, if the center-size framework can accurately model Existential Positive First-order (EPFO) logical queries, the dimensionality of the logical query embeddings would need to be $\mathcal{O}(|\mathcal{V}|)$~\cite{Query2box}, where $|\mathcal{V}|$ is the number of entities, which is not low-dimensional.
The low-level object and high-dimensional embedding are key limitations that prevent the center-size framework from being able to represent all logical operations.
Secondly, \textbf{the coupling of feature description and logical representation in the center-size framework makes it difficult to generalize to new features.}.
In real world KGs, the features of entities are usually heterogeneous, such as text, image, and graph.
These features contribute to reasoning, so a good reasoning framework should consider the extension to integrate new features.
However, the current center-size framework has a one-to-one constraint on the center and size, which limits the extensibility and flexibility of the model.
For example, ConE requires that one axis must correspond to one aperture, to represent sector-cone.
Adding a new feature would require a second axis, but it would be difficult for ConE to generalize to the case where two axes correspond to one aperture, because this would not be a valid parameterization of a sector-cone.

Therefore, it is of great significance to propose a new embedding-based framework that (1) keeps closed under all logical operations and (2) is easy to generalize to new features.
We consider dividing the embeddings of entities and queries into two loosely coupled parts, feature part and logic part, and then learns each part independently.
In this way, we can easily extend the model to new features (by adding new feature parts) and keep all logical operations closed (by introducing closed fuzzy logic).
For the feature parts, neural networks are adopted to capture the query representation and logical information.
For the logic parts, we propose to use \textbf{vector logic} to guarantee to model at least all logical operators, directly.
We use vector logic because it is a fuzzy logic system based on matrix algebra, easy to cope with neural networks.

Feature-Logic framework has the following advantages:
(1) \xhdr{All neural logical operators are closed}
The FOL operations include conjunction, disjunction, negation, etc.
These operations on feature-logic embeddings generate feature-logic embedding, which means that all neural operators are closed.
Furthermore, we also prove that these operators based on feature-logic embedding obey the rules of real logical operations in Section~\ref{sec:model:theoretical_analysis}.
(2) \xhdr{High extensibility and flexibility}
Unlike centers-size framework, our framework has fewer constraints on each part.
On the one hand, feature-logic embedding can extend to multiple independent feature parts, which help to integrate heterogeneous information as a result of more precise description of entities in KGs.
The feature part is also able to choose a variety of feature spaces, such as limited or infinite real space (and complex space, theoretically).
On the other hand, the logic part can be modified without impacting on the feature part.
It is also allowed to replace vector logic with other fuzzy system on the logic part.

We summarized our contributions as follows:
(1) We propose a closed embedding-based framework FLEX for knowledge graph reasoning, which not only supports various feature spaces, but also has all closed logical operators, including conjunction, disjunction, negation, etc.
(2) To our best knowledge, we are the first to introduce vector logic into the field of logical reasoning over knowledge graphs.
(3) Experiments on benchmark datasets demonstrate that our framework significantly outperforms state-of-the-art models. The source code is also available online (Appendix~\ref{sec:appendix:train}).

\section{Related Work}\label{sec:relatedwork}





To solve Knowledge Graph Reasoning (KGR) problem, embedding-based methods represent the query set as a vector in low-dimensional spaces, where vectors of answer entities and queries are close to each other.
Plenty of models are proposed to create the embeddings, such as (1) probability distribution (2) geometric object (3) fuzzy logic and (4) others~\cite{EmQL, QuantumE, QuantumE+}.

\textbf{Probability Distribution} is a natural choice for modeling uncertainty of query set.
GaussianPath~\cite{GaussianPath} uses Gaussian distribution for multi-hop reasoning. But it doesn't take logical computation into account.
BetaE~\cite{BetaE} and PERM~\cite{PERM} are two representative models that use Beta distribution and mixture of Gaussian distributions to model the uncertainty of answers.
BetaE supports projection, intersection and negation, while PERM can handle projection, intersection and union.
However, BetaE must use negation and intersection to represent difference. But since the difference of Beta distributions is not a Beta distribution, the family of Beta distribution is not closed for any logical operators.
PERM ignores negation because the complement of Gaussian distribution is not a Gaussian distribution.
In comparison, our framework is closed under all FOL operations.
The probabilistic interpretation of vector logic in our framework can also represent uncertainty.

\textbf{Geometric Object} is an intuitive way to model the query set.
These methods use point~\cite{GQE}, box~\cite{Query2box, NewLook}, sector-cone~\cite{ConE}, hyperboloids in a Poincaré ball~\cite{HypE} and so on to represent the query set.
They connect geometric object with set transformation, but their logical representation is limited by the embedding dimension~\cite{Query2box}.
In addition, the parameterization of geometric objects is strict, which makes it difficult to generalize to new features.
For example, ConE~\cite{ConE} can handle union using DNF. But since the union of sector-cones is a cone instead of sector-cone, the family of sector-cone is not closed under union.

\textbf{Fuzzy Logic} methods regard the query set as a fuzzy set, where reasoning is performed by fuzzy operations.
CQD~\cite{CQD} proposes to train a projection operator (\eg TransE\cite{TransE}, RotatE\cite{RotatE}, ComplEx\cite{ComplEx}, etc.) and then uses t-norm based fuzzy logic system to rank entities.
The performance is impressive, but in the paper of CQD, it ignores negation, which is the atomic operation in FOL operations.
FuzzQE~\cite{FuzzQE} and LogicE~\cite{LogicE} consider product logic, Gödel logic and Łukasiewicz logic~\cite{KlementTNormBook}, which are most prominent fuzzy logics.
These methods are not embedding-based, because they use fuzzy set rather than embedding to represent the query set.
In this paper, we introduce another fuzzy logic, namely vector logic, which is easy to cope with neural networks.
Our framework is not only an embedding-based model, but also a fuzzy logic system with all well-designed neural logical operators.





\section{Method}\label{sec:method}
In this section, we propose feature-logic embedding framework (FLEX) for KGR.
We first introduce how to reason using first-order logic in Section~\ref{sec:model:common} and vector logic in Section~\ref{sec:model:vector_logic}
Afterwards, we present feature-logic embeddings and the methods to learn in Section~\ref{sec:model:component}, \ref{sec:model:operators} and \ref{sec:model:learn_embedding}.
Furthermore, we provide theoretical analysis in Section~\ref{sec:model:theoretical_analysis} to show that FLEX is closed, and obeys the rules of real logical operations.

\subsection{First-Order Logic for Knowledge Graph Reasoning}
\label{sec:model:common}

\xhdr{Knowledge Graph (KG)} $\mathcal{G} = \{\mathcal{V}, \mathcal{R}, \mathcal{T}\}$ consists of entity set $\mathcal{V}$, relation set $\mathcal{R}$ and triple set $\mathcal{T} = \{(h, r, t)| h, t\in \mathcal{V}; r\in \mathcal{R}\}$, where $h$, $r$, $t$ denotes head entity, relation, tail entity respectively.

\xhdr{First-Order Logic (FOL)} In the query embedding literature, FOL queries typically involve logical operations such as existential quantification ($\exists$), monadic operations (identity, negation $\neg$) and dyadic operations (conjunction $\cap$, disjunction $\cup$, implication $\rightarrow $, equivalence $\leftrightarrow $, exclusive or $\oplus$, etc.). However, universal quantification ($\forall$) is not typically included, as it is not typically applicable in real-world knowledge graphs where no entity connects with all other entities~\cite{Query2box,BetaE,ConE}.


To formulate FOL query $q$, we use its \underline{d}isjunctive \underline{n}ormal \underline{f}orm (DNF)\cite{Query2box} which is defined as:
\begin{equation}
  q = V_?, \exists V_1, \cdots, V_k:c_1 \lor c_2 \lor \cdots \lor c_n
\end{equation}
where $V_?$ is target variable, $V_1, \cdots, V_k$ are existentially quantified bound variables, $c_i$ represents conjunctions, \ie, $c_i = e_{i1} \land e_{i2} \land \cdots \land e_{im}$ and literal $e$ represents an atomic formula or its negation, \ie, $e_{ij} = \mathcal{P}_{r}(v_a, V)$ or $\lnot \mathcal{P}_{r}(v_a, V)$ or $\mathcal{P}_{r}(V', V)$ or $\lnot \mathcal{P}_{r}(V', V)$, where $v_a \in V_a \subseteq \mathcal{V}$ is a non-variable anchor entity set, $V \in \{V_?, V_1, \cdots, V_k\}$, $V'\in \{ V_1, \cdots, V_k\}$ and $V \ne V'$, $\mathcal{P}_{r} : \mathcal{V} \times \mathcal{V} \to \{\text{True}, \text{False}\}$ is the predicate corresponding to relation $r\in\mathcal{R}$. $\mathcal{P}_{r}(V_h, V_t)=\text{True}$ if and only if $(h, r, t)$ is a factual triple in $\mathcal{T}$, $\forall h \in V_h \subseteq \mathcal{V}, t\in V_t \subseteq \mathcal{V}$.
Answering query $q$ is to find the entity set $\llbracket q\rrbracket \subset \mathcal{V}$, where $v\in\llbracket q\rrbracket$ if and only if $q[v]$ is true.


\xhdr{Computation Graph}
The reasoning procedure for FOL query can be represented as a computation graph (Figure~\ref{fig:query}), where nodes represent entity sets and edges represent logical operations.
To make the computation work in embedding space, query embedding (QE) methods generate low-dimensional embeddings for nodes and map edges to logical operators as follows:
(1) \underline{Projection Operator $\mathcal{P}$}: Given an entity set $S \subseteq \mathcal{V}$ and a relation $r \in \mathcal{R}$, project operation maps $S$ to tail entity set $S'=\cup_{v\in S}\{v'|(v, r,v')\in\mathcal{T}\}$.
(2) \underline{Monadic Operators}: \textbf{Identity} and \textbf{Negation} $\mathcal{N}$. The complement set of given entity set $S$ in KG is $\bar{S} = \mathcal{V} \backslash S$.
(3) \underline{Dyadic Operators}: \textbf{Intersection} $\mathcal{I}$, \textbf{Union} $\mathcal{U}$, \textbf{Implication} $\mathcal{L}$, etc. Given a set of entity sets $\{S_1, \cdots, S_n\}$, the intersection (union, etc.) operation computes logical intersection (union, etc.) of these sets $\cap_{i=1}^n S_i$ ($\cup_{i=1}^n S_i$, etc.).



\subsection{Vector Logic}
\label{sec:model:vector_logic}

We follow the definition of vector logic in \cite{vector_logic}.
Vector logic enables us to use fuzzy logic in embedding space.
Vector logic is an elementary logical model based on matrix algebra.
In vector logic, true value are mapped to the vector, and logical operators are executed by matrix computation.

\xhdr{Truth Value Vector Space $\mathbb{V}$}
A two-valued vector logic use two $d$-dimensional ($d \ge 2$) column vectors $\vec{s}$ and $\vec{n}$ to represent true and false in the classic binary logic.
$\vec{s}$ and $\vec{n}$ are real-valued, normally orthogonal to each other, and normalized vectors, \ie, $\|\vec{s}\|=1, \|\vec{n}\|=1, \vec{s}^T\vec{u}=0$.

\xhdr{Operators}
The basic logical operators are associated with its own matrices by vectors in truth value vector space.
Two common types of operators are monadic and dyadic.

(1) Monadic Operators:
Monadic operators are functions: $\mathbb{V} \to \mathbb{V}$.
Two examples are Identity $I = \vec{s} \vec{s}^T + \vec{n} \vec{n}^T$ and Negation $N = \vec{n} \vec{s}^T + \vec{s} \vec{n}^T$ such that $I\vec{s} = \vec{s}, I\vec{n} = \vec{n}, N\vec{n} = \vec{s}, N\vec{s} = \vec{n}$.

(2) Dyadic Operators:
Dyadic operators are functions: $\mathbb{V} \otimes \mathbb{V} \to \mathbb{V}$, where $\otimes$ denotes Kronecker product.
Dyadic operators include conjunction $C$, disjunction $D$, implication $L$, exclusive or $X$, etc.
For example, the conjunction between two logical propositions $(p\land q)$ is performed by the matrix multiplication of $C$ and $\vec{u}\otimes \vec{v}$, \ie, $C(\vec{u}\otimes \vec{v})$, where $C=\vec{s}(\vec{s} \otimes \vec{s})^T+\vec{n}(\vec{s} \otimes \vec{n})^T+\vec{n}(\vec{n} \otimes \vec{s})^T+\vec{n}(\vec{n} \otimes \vec{n})^T$, $\vec{u},\vec{v}$ are any two vectors in truth value vector space.
It can be verified that $C(\vec{s} \otimes \vec{s})=\vec{s},C(\vec{s} \otimes \vec{n}) = C(\vec{n} \otimes \vec{s}) = C(\vec{n} \otimes \vec{n}) = \vec{n}$, which satisfies the true value table of conjunction.

\xhdr{Many-valued Two-dimensional Logic}
Many-valued logic are introduced to include uncertainties in the logic vector.
Weighting $\vec{s}$ and $\vec{n}$ by probabilities, uncertainties are introduced: $\vec f = \epsilon \vec{s} + (1-\epsilon) \vec{n}$, where $\epsilon \in [0,1]$.
Besides, operations on vectors can be simplified to computation on the scalar of these vectors.
For example, given two vectors $\vec{u} = \alpha \vec{s} + (1-\alpha) \vec{n}, \vec{v}= \alpha' \vec{s} + (1-\alpha') \vec{n}$, we have:
\begin{equation}\label{eq:vector_logic}
  \begin{aligned}
    \text{NOT}(\vec{u})           = & N\vec{u} = (1 - \alpha)\vec{s}+\alpha\vec{n} & \text{NOT}(\alpha)           & =\vec{s}^T N\vec{u} = 1 - \alpha                                    \\
    \text{OR}(\vec{u}, \vec{v})   = & D(\vec{u} \otimes \vec{v})                   & \text{OR}(\alpha, \alpha')   & =\vec{s}^T D(\vec{u} \otimes \vec{v})=\alpha+\alpha'-\alpha\alpha'  \\
    \text{AND}(\vec{u}, \vec{v})  = & C(\vec{u} \otimes \vec{v})                   & \text{AND}(\alpha, \alpha')  & =\vec{s}^T C(\vec{u} \otimes \vec{v})=\alpha\alpha'                 \\
    \text{IMPL}(\vec{u}, \vec{v}) = & L(\vec{u} \otimes \vec{v})                   & \text{IMPL}(\alpha, \alpha') & =\vec{s}^T L(\vec{u} \otimes \vec{v})=1-\alpha(1-\alpha')           \\
    \text{XOR}(\vec{u}, \vec{v})  = & X(\vec{u} \otimes \vec{v})                   & \text{XOR}(\alpha, \alpha')  & =\vec{s}^T X(\vec{u} \otimes \vec{v})=\alpha+\alpha'-2\alpha\alpha'
  \end{aligned}
\end{equation}

\subsection{Feature-Logic Embeddings for Queries and Entities}
\label{sec:model:component}

In this section, we design embeddings for queries and entities.
To model the logical information hidden in the query embedding, we propose to consider a part of the embedding as logic part, while the rest as feature part.
The logic part obeys the rules of real logical operations in vector logic and doesn't depend on the feature part.
Formally, the embedding of $\llbracket q\rrbracket$ is $\mathbf{V}_q=(\boldsymbol{\theta}_{f}, \boldsymbol{\theta}_{l})$ where $\boldsymbol{\theta}_{f}\in [-L,L]^d$ is the feature part and $\boldsymbol{\theta}_{l} \in [0,1]^d$ is the logic part, $d$ is the embedding dimension and $L>0$ is a boundary.
In vector logic, the parameter $\boldsymbol{\theta}_{l}$ is the uncertainty $\boldsymbol{\theta}_{l}\vec{s} + (1-\boldsymbol{\theta}_{l})\vec{n}$ of the feature.
An entity $v\in\mathcal{V}$ is represented as a single-element query set without uncertainty.
We propose to represent an entity as the query with logic part $\boldsymbol{0}$, which indicates that the entity's uncertainty is $0$. Formally, the embedding of entity $v$ is $\textbf{v}=(\boldsymbol{\theta}_{f}, \boldsymbol{0})$, where $\boldsymbol{\theta}_{f}\in [-L,L]^d$ is the feature part and $\boldsymbol{0}$ is a $d$-dimensional vector with all elements being 0.

\subsection{Logical Operators for Feature-Logic Embeddings}
\label{sec:model:operators}

\begin{figure*}[t]
  \vspace{-3mm}
  \centering
  \begin{subfigure}[b]{0.24\textwidth}
    \includegraphics[width=\textwidth]{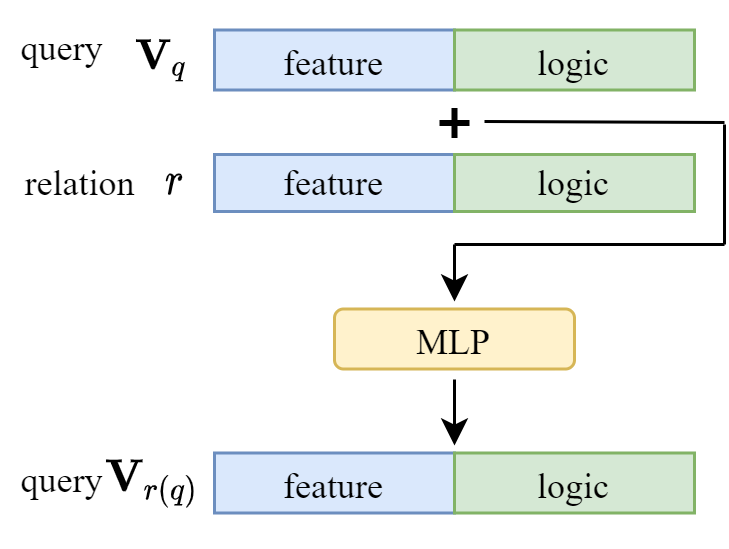}
    \vspace{-5mm}
    \caption{Projection}
    \label{subfig:projection}
  \end{subfigure}\hfil
  \begin{subfigure}[b]{0.23\textwidth}
    \includegraphics[width=\textwidth]{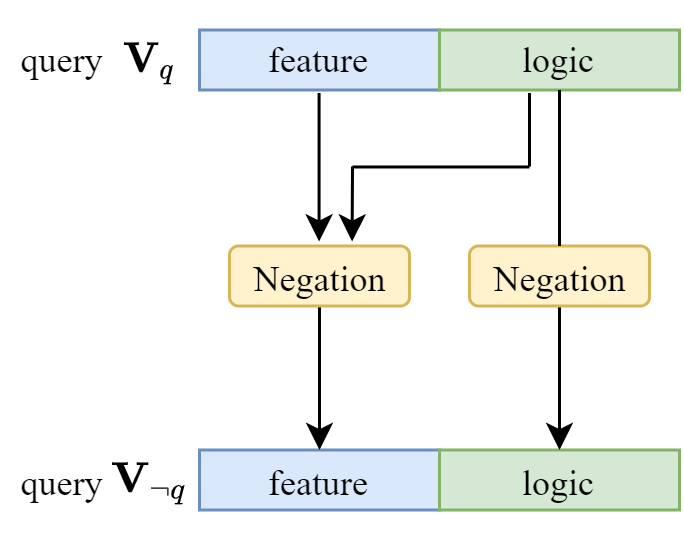}
    \vspace{-5mm}
    \caption{Negation}
    \label{subfig:complement}
  \end{subfigure}\hfil
  \begin{subfigure}[b]{0.265\textwidth}
    \includegraphics[width=\textwidth]{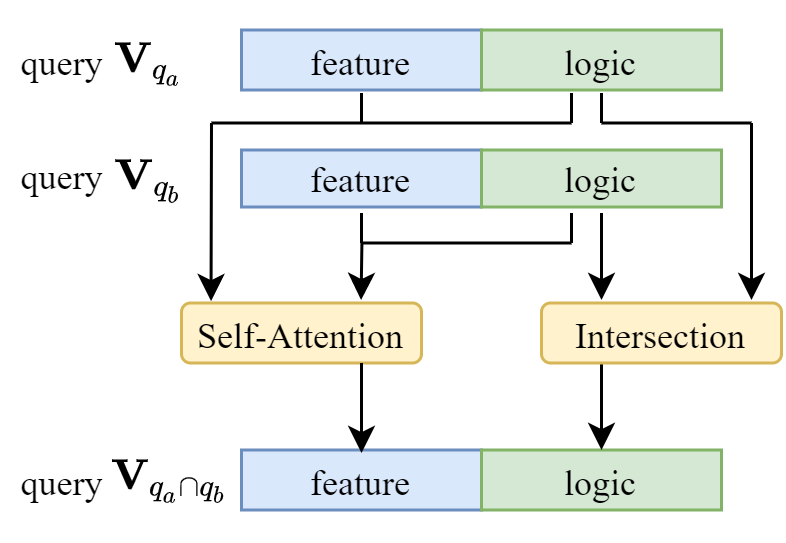}
    \vspace{-5mm}
    \caption{Intersection}
    \label{subfig:intersection}
  \end{subfigure}\hfil
  \begin{subfigure}[b]{0.265\textwidth}
    \includegraphics[width=\textwidth]{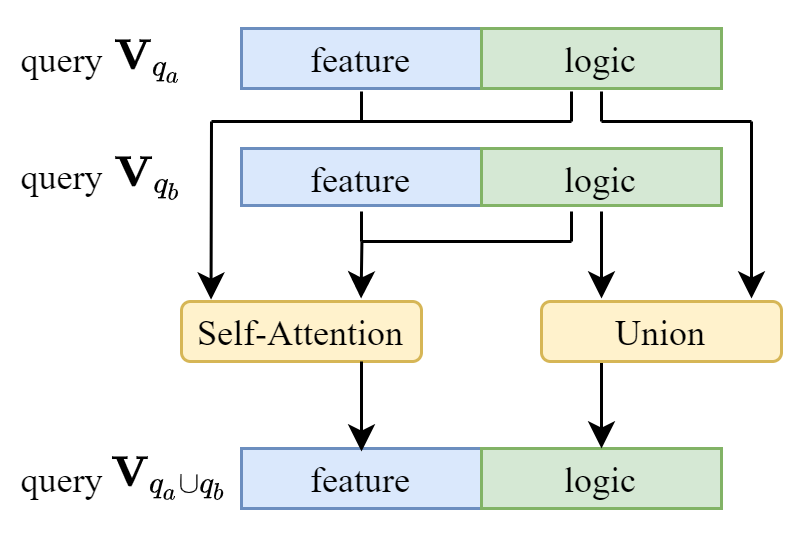}
    \vspace{-5mm}
    \caption{Union}
    \label{subfig:union}
  \end{subfigure}\hfil
  \vskip -1mm
  \caption{FLEX's neural logical operators. Each query embedding consists of feature part and logic part. The feature part depends on previous feature and logic parts. The logic part obeys the rules of real logical operations independently.}
  \label{fig:models}
  \vspace{-4mm}
\end{figure*}

In this section, we introduce the designed logical operators, including projection, intersection, complement, union, and other dyadic operators.


\xhdr{Projection Operator $\mathcal{P}$}
The goal of operator $\mathcal{P}$ is to map an entity set to another entity set under a given relation. We define a function $f_r : \mathbf{V}_q \mapsto \mathbf{V}_q'$ in the embedding space to represent $\mathcal{P}$. To implement $f_r$, we first represent relations as translations on query embeddings and assign each relation with relational embedding $\mathbf{r}=(\boldsymbol{\theta}_{f,r}, \boldsymbol{\theta}_{l,r})$. Then $f_r$ is defined as:
\begin{equation}
  f_r(\mathbf{V}_q) = g(\textbf{MLP}([\boldsymbol{\theta}_{f} + \boldsymbol{\theta}_{f,r}; \boldsymbol{\theta}_{l} + \boldsymbol{\theta}_{l,r}]))
\end{equation}
where $\textbf{MLP} : \mathbb{R}^{2d} \to \mathbb{R}^{2d} $ is a multi-layer perceptron network (MLP), $[\cdot;\cdot]$ is concatenation and $g$ is an activate function to generate $\boldsymbol{\theta}_{f}'\in [-L, L]^d, \boldsymbol{\theta}_{l}'\in[0, 1]^d$.
We define $g$ as (Figure~\ref{subfig:projection}):
\begin{equation}
  g(\mathbf{x})=[L\tanh(\mathbf{x}[0:d]);\;\sigmoid(\mathbf{x}[d:2d])]
\end{equation}
where $\mathbf{x}[0:d]$ is the slice containing element $x_i$ of vector $\mathbf{x}$ with index $0\leq i < d$, $\tanh(\cdot)$ is hyperbolic tangent function and $\sigmoid(\cdot)$ is sigmoid function.

\xhdr{Negation Operator $\mathcal{N}$}
The aim of $\mathcal{N}$ is to identify the complement of query set $\llbracket q\rrbracket$ such that $\llbracket \lnot q\rrbracket=\mathcal{V}\backslash \llbracket q\rrbracket$. Suppose that $\mathbf{V}_q=(\boldsymbol{\theta}_{f}, \boldsymbol{\theta}_{l})$ and $\mathbf{V}_{\lnot q}=(\boldsymbol{\theta}_{f}', \boldsymbol{\theta}_{l}')$.
$\mathcal{N}$ is defined as (Figure~\ref{subfig:complement}):
\begin{equation}
    \boldsymbol{\theta}_{f}' = L\tanh (\textbf{MLP}([\boldsymbol{\theta}_{f};\boldsymbol{\theta}_{l}])), \;\;\;\;\;\;
    \boldsymbol{\theta}_{l}' = 1 - \boldsymbol{\theta}_{l}
\end{equation}
where $\textbf{MLP}:\mathbb{R}^{2d} \to \mathbb{R}^{d}$ is a MLP network, $[\cdot;\cdot]$ is concatenation.
The computation of logic part follows NOT operation in vector logic (Eq.~\ref{eq:vector_logic}).

\xhdr{Dyadic Operators}
\label{sec:model:extension}
Benefit from vector logic, our framework is able to directly model all dyadic operators.
We design similar structure for all dyadic operators.
Below we take the Intersection $\mathcal{I}$ and Union $\mathcal{U}$ for example (Figure~\ref{subfig:intersection}~\ref{subfig:union}).
The goal of operator $\mathcal{I}$ ($\mathcal{U}$) is to represent $\llbracket q\rrbracket = \cap_{i=1}^n \llbracket q_i \rrbracket$ ($\llbracket q\rrbracket = \cup_{i=1}^n \llbracket q_i \rrbracket$).
Suppose that $\mathbf{V}_q=(\boldsymbol{\theta}_{f}, \boldsymbol{\theta}_{l})$ and $\mathbf{V}_{q_i}=(\boldsymbol{\theta}_{q_i,f}, \boldsymbol{\theta}_{q_i,l})$ are feature-logic embeddings for $\llbracket q\rrbracket$ and $\llbracket q_i\rrbracket$, respectively.

First, the feature part outputted from operator is given by the self-attention neural network:
\begin{equation}
    \boldsymbol{\theta}_{f}' =\sum_{i=1}^n \boldsymbol{\alpha}_i \boldsymbol{\theta}_{q_i,f},  \;\;\;\;\;\;
    \boldsymbol{\alpha}_i = \frac{\exp (\textbf{MLP}([\boldsymbol{\theta}_{q_i,f};\boldsymbol{\theta}_{q_i,l}]))}{\sum_{j=1}^n \exp (\textbf{MLP}([\boldsymbol{\theta}_{q_j,f};\boldsymbol{\theta}_{q_j,l}]))},\;\;\;\;\;\;
    \textbf{MLP}:\mathbb{R}^{2d} \to \mathbb{R}^{d}
\end{equation}
where $\boldsymbol{\alpha}_i$ is the attention weight to notice the changes of logic part, $[\cdot;\cdot]$ is concatenation.

Second, the logic part of $\mathcal{I}$ ($\mathcal{U}$) should follow \textbf{AND} (\textbf{OR}) operation in vector logic.
Therefore, $\mathcal{I}$ uses $\boldsymbol{\theta}_{l}' = \prod_{i=1}^n \boldsymbol{\theta}_{q_i,l}$  and $\mathcal{U}$ has $\boldsymbol{\theta}_{l}' =\sum_{i=1}^n \boldsymbol{\theta}_{q_i,l}
-\sum_{1\leqslant i < j\leqslant n} \boldsymbol{\theta}_{q_i,l} \boldsymbol{\theta}_{q_j,l}
+\sum_{1\leqslant i < j < k\leqslant n} \boldsymbol{\theta}_{q_i,l} \boldsymbol{\theta}_{q_j,l} \boldsymbol{\theta}_{q_k,l}
+\dotsc +(-1)^{n-1} \prod_{i=1}^n \boldsymbol{\theta}_{q_i,l}$.
When $n=2$, the logic part is $\boldsymbol{\theta}_{l}' = \boldsymbol{\theta}_{q_1,l}\boldsymbol{\theta}_{q_2,l}$ and $\boldsymbol{\theta}_{l}' =\boldsymbol{\theta}_{q_1,l} + \boldsymbol{\theta}_{q_2,l} - \boldsymbol{\theta}_{q_1,l}\boldsymbol{\theta}_{q_2,l}$, corresponding to the \textbf{AND} and \textbf{OR} operation in vector logic (Eq.~\ref{eq:vector_logic}). The computation complexity is $O(nd)$ where $n$ denotes the number of queries to union and $d$ is the embedding dimension.
Note that the logic part $\boldsymbol{\theta}_{l}'$ only depends on previous logic parts, according to vector logic. The self-attention neural network will learn the hidden information from logic and leverage to the feature part. This architecture makes flexibility come true, different from center-size framework.

In fuzzy logic, it is common to use $\min(\cdot)$ and $\max(\cdot)$ as \textbf{AND} and \textbf{OR} operations respectively.
Therefore, we also propose another intersection operator $\mathcal{I}_2$, whose feature part is the same as $\mathcal{I}$ but the logic part is $\boldsymbol{\theta}_{l}' =\min_{i=1}^{n}\{\boldsymbol{\theta}_{q_i,l}\}$.
The other union operator $\mathcal{U}_2$ has $\boldsymbol{\theta}_{l}' =\max_{i=1}^{n}\{\boldsymbol{\theta}_{q_i,l}\}$.

\subsection{Learning Feature-Logic Embeddings}
\label{sec:model:learn_embedding}

We expect to achieve high scores for the answers $v\in \llbracket q\rrbracket$ to query $q$,  and low scores for $v'\notin \llbracket q\rrbracket$.
Therefore, we firstly define a distance function to measure the distance between a given entity embedding and a query embedding, and then we train the model with negative sampling loss.

\xhdr{Distance Function} Given an entity embedding $\mathbf{v}=(\boldsymbol{\theta}_{v,f}, \boldsymbol{0})$ and a query embedding $\mathbf{V}_q=(\boldsymbol{\theta}_{q,f}, \boldsymbol{\theta}_{q,l})$, we define the distance $d$ between this entity $v$ and the query $q$ as the sum of the feature distance (between feature parts) and the logic part (to expect uncertainty to be 0):  $d(\mathbf{v};\mathbf{V}_q) = \|\boldsymbol{\theta}_{v,f} - \boldsymbol{\theta}_{q,f}\|_1 +\boldsymbol{\theta}_{q,l}$, where $\|\cdot\|_1$ is the $L_1$ norm and $+$ is the element-wise addition.
However, for cases where there is no disjunctive query for training, because the union operator cannot be trained, we use DNF technique~\cite{BetaE} to transform all queries to DNF.
Given a DNF query $q=q_1\lor \dots\lor q_n$, we instead use DNF distance $d_{\text{DNF}}(\textbf{v};\textbf{V}_{q})  =\textbf{OR}\{d(\textbf{v};\textbf{V}_{q_1}),\dots,d(\textbf{v};\textbf{V}_{q_n})\}$ where $\textbf{OR}$ is OR operation in vector logic (Eq.~\ref{eq:vector_logic}).

\xhdr{Loss Function} Given a training set of queries, we optimize a negative sampling loss $L=-\log\sigma(\gamma-d(\mathbf{v};\mathbf{V}_q))-\frac{1}{k}\sum_{i=1}^k\log\sigma(d(\mathbf{v}'_i;\mathbf{V}_q)-\gamma)$ where $\gamma>0$ is a fixed margin, $v\in\llbracket q\rrbracket$ is a positive entity, $v_i'\notin \llbracket q \rrbracket$ is the $i$-th negative entity, $k$ is the number of negative entities, and $\sigma(\cdot)$ is the sigmoid function.

\subsection{Theoretical Analysis}
\label{sec:model:theoretical_analysis}

We have the following propositions with proofs in Appendix~\ref{sec:appendix:theoretical_analysis}:

\begin{proposition}
  \label{prop:Closure}
  \textbf{Closure}: The projection operator, monadic and dyadic operators are closed under feature-logic embeddings of queries and entities.
\end{proposition}

\begin{proposition}
  \label{prop:Commutativity}
  \textbf{Commutativity}: Given feature-logic embedding $\mathcal{V}_{q_a}, \mathcal{V}_{q_b}$, we have $\mathcal{I}(\mathcal{V}_{q_a}, \mathcal{V}_{q_b}) = \mathcal{I}(\mathcal{V}_{q_b}, \mathcal{V}_{q_a})$ and $\mathcal{U}(\mathcal{V}_{q_a}, \mathcal{V}_{q_b}) = \mathcal{U}(\mathcal{V}_{q_b}, \mathcal{V}_{q_a})$, $\mathcal{I}_2(\mathcal{V}_{q_a}, \mathcal{V}_{q_b}) = \mathcal{I}_2(\mathcal{V}_{q_b}, \mathcal{V}_{q_a})$ and $\mathcal{U}_2(\mathcal{V}_{q_a}, \mathcal{V}_{q_b}) = \mathcal{U}_2(\mathcal{V}_{q_b}, \mathcal{V}_{q_a})$.
\end{proposition}


\begin{proposition}
  \label{prop:Idempotence}
  \textbf{Idempotence}: Given feature-logic embedding $\mathcal{V}_q$, we have $\mathcal{I}_2(\{\mathcal{V}_q,\mathcal{V}_q,\cdots,\mathcal{V}_q\}) = \mathcal{V}_q$ and $\mathcal{U}_2(\{\mathcal{V}_q,\mathcal{V}_q,\cdots,\mathcal{V}_q\}) = \mathcal{V}_q$.
\end{proposition}




The Proposition~\ref{prop:Closure} indicates FLEX is closed. The Propositions~\ref{prop:Commutativity},\ref{prop:Idempotence} show that our designed intersection and union operators obey the rules of real logical operations.

\section{Experiments}\label{sec:exp}

\subsection{Experimental Settings}
\label{sec:exp:setting}


\xhdr{Dataset}
\label{sec:exp:dataset}
We use three datasets: FB15k~\cite{Freebase}, FB15k-237 (FB237)~\cite{FB237} and NELL995 (NELL)~\cite{NELL}.
For fair comparison, we use the same query structures in~\cite{BetaE}.
The training and validation queries consist of five conjunctive structures ($1p/2p/3p/2i/3i$) and five structures with negation ($2in/3in/inp/pni/pin$).
We also evaluate the ability of generalization, \ie answering queries with structures that models have never seen during training.
The extra query structures include $ip/pi/2u/up$.
Please refer to Appendix~\ref{sec:appendix:datasets} for more details about datasets.


\xhdr{Evaluation}
\label{sec:exp:eval}
We adopt the same evaluation protocol as that in BetaE~\cite{BetaE}. Firstly, we build three KGs: the training KG (training edges), the validation KG (training + validation edges) and the test KG (training + validation + test edges). Given a test query $q$, for each non-trivial answer $v\in \llbracket q\rrbracket_{\text{test}} - \llbracket q\rrbracket_{\text{valid}} $ of the query $q$, we rank it against non-answer entities $\mathcal{V} - \llbracket q\rrbracket_{\text{test}}$. Then we calculate Hits@k (k=1,3,10) and Mean Reciprocal Rank (MRR) based on the rank. Please refer to Appendix~\ref{sec:appendix:eval} for the definition of Hits@k and MRR. For all metrics, the higher, the better.

\xhdr{Baselines}
\label{sec:exp:baselines}
We compare FLEX against four state-of-the-art models, including GQE~\cite{GQE}, Query2Box (Q2B)\cite{Query2box}, BetaE~\cite{BetaE} and ConE~\cite{ConE}.

\subsection{Main Results}
\label{sec:exp:result}

\xhdr{Queries without Negation.} Table~\ref{table:main_results_without_negation} shows the results on queries without negation (EPFO queries).
Overall, FLEX outperforms compared models.
FLEX achieves on average 5.2\%, 5.1\% and 3.7\% relative improvement MRR over previous state-of-the-art (SOTA) ConE on FB15k, FB237 and NELL respectively, which demonstrates the advantages of feature-logic framework. 
Besides, FLEX gains an impressive improvement on queries \textbf{1p}.
For example, FLEX outperforms ConE by 8.9\% for \textbf{1p} query on NELL.
The good results on \textbf{1p} queries enable our further comparison on traditional link prediction tasks with SOTA models, such as TuckER~\cite{TuckER}.
Please refer to Appendix~\ref{LP and MUL}.

\xhdr{Queries with Negation.} Table~\ref{table:main_results_with_negation} shows the results on queries with negation.
Generally speaking, FLEX outperforms BetaE and ConE.
For relative MRR improvement compared with ConE, FLEX achieves 4.1\% on FB15k, 3.4\% on FB237 and 4.7\% on NELL.
We attribute the improvement to that our negation operator is also neural, easy to cooperate with other neural logical operators, and has a good generalization ability.
The improvement demonstrates the effectiveness of our method.

\begin{table*}[!t]
  \caption{MRR results for answering queries without negation ($\exists$, $\land$, $\lor$) on FB237 and NELL. The best results are in bold. \textbf{AVG} denotes average performance.}
  \label{table:main_results_without_negation}
  \centering
  \resizebox{0.8\columnwidth}{!}{
    \begin{tabular}{c c c c c c c c c c c c c c }
      \toprule
      \textbf{Dataset}       & \textbf{Model} & \textbf{1p}   & \textbf{2p}   & \textbf{3p}   & \textbf{2i}   & \textbf{3i}   & \textbf{pi}   & \textbf{ip}   & \textbf{2u}   & \textbf{up}   & \textbf{AVG}  \\
      \midrule
      \multirow{5}{*}{FB15k} & GQE            & 53.9          & 15.5          & 11.1          & 40.2          & 52.4          & 27.5          & 19.4          & 22.3          & 11.7          & 28.2          \\
                             & Q2B            & 70.5          & 23.0          & 15.1          & 61.2          & 71.8          & 41.8          & 28.7          & 37.7          & 19.0          & 40.1          \\
                             & BetaE          & 65.1          & 25.7          & 24.7          & 55.8          & 66.5          & 43.9          & 28.1          & 40.1          & 25.2          & 41.6          \\
                             & ConE           & 73.3          & 33.8          & 29.2          & 64.4          & 73.7          & 50.9          & 35.7          & \textbf{55.7}          & 31.4          & 49.8          \\
                             & FLEX           & \textbf{77.1} & \textbf{37.4} & \textbf{31.6} & \textbf{66.4} & \textbf{75.2} & \textbf{54.2} & \textbf{42.4} & 52.9 & \textbf{34.3} & \textbf{52.4} \\
      \midrule
      \multirow{5}{*}{FB237} & GQE            & 35.2          & 7.4           & 5.5           & 23.6          & 35.7          & 16.7          & 10.9          & 8.4           & 5.8           & 16.6          \\
                             & Q2B            & 41.3          & 9.9           & 7.2           & 31.1          & 45.4          & 21.9          & 13.3          & 11.9          & 8.1           & 21.1          \\
                             & BetaE          & 39.0          & 10.9          & 10.0          & 28.8          & 42.5          & 22.4          & 12.6          & 12.4          & 9.7           & 20.9          \\
                             & ConE           & 41.8          & 12.8          & 11.0          & 32.6          & 47.3          & 25.5          & 14.0          & 14.5          & 10.8          & 23.4          \\
                             & FLEX           & \textbf{43.6} & \textbf{13.1} & \textbf{11.1} & \textbf{34.9} & \textbf{48.4} & \textbf{27.4} & \textbf{16.1} & \textbf{15.4} & \textbf{11.1} & \textbf{24.6} \\
      \midrule
      \multirow{5}{*}{NELL}  & GQE            & 33.1          & 12.1          & 9.9           & 27.3          & 35.1          & 18.5          & 14.5          & 8.5           & 9.0           & 18.7          \\
                             & Q2B            & 42.7          & 14.5          & 11.7          & 34.7          & 45.8          & 23.2          & 17.4          & 12.0          & 10.7          & 23.6          \\
                             & BetaE          & 53.0          & 13.0          & 11.4          & 37.6          & 47.5          & 24.1          & 14.3          & 12.2          & 8.5           & 24.6          \\
                             & ConE           & 53.1          & 16.1          & 13.9          & 40.0          & \textbf{50.8} & 26.3          & 17.5          & 15.3          & 11.3          & 27.2          \\
                             & FLEX           & \textbf{57.8} & \textbf{16.8} & \textbf{14.7} & \textbf{40.5} & \textbf{50.8} & \textbf{27.3} & \textbf{19.4} & \textbf{15.6} & \textbf{11.6} & \textbf{28.2} \\
      \bottomrule
    \end{tabular}
  }
  \vspace{-3mm}
\end{table*}

\begin{table*}[!t]
  \caption{MRR results for answering queries with negation on FB237, and NELL. The best results are in bold. \textbf{AVG} denotes average performance.}
  \label{table:main_results_with_negation}
  \centering
  \resizebox{0.55\columnwidth}{!}{
    \begin{tabular}{ c c  c c c c  c c c c }
      \toprule
      \textbf{Dataset}       & \textbf{Model} & \textbf{2in}  & \textbf{3in}  & \textbf{inp}  & \textbf{pin}  & \textbf{pni}  & \textbf{AVG}  \\
      \midrule
      \multirow{3}{*}{FB15k} & BetaE          & 14.3          & 14.7          & 11.5          & 6.5           & 12.4          & 11.8          \\
                             & ConE           & 17.9          & 18.7          & 12.5          & 9.8           & 15.1          & 14.8          \\
                             & FLEX           & \textbf{18.0} & \textbf{19.3} & \textbf{14.2} & \textbf{10.1} & \textbf{15.2} & \textbf{15.4} \\
      \midrule
      \multirow{3}{*}{FB237} & BetaE          & 5.1           & 7.9           & 7.4           & 3.6           & 3.4           & 5.4           \\
                             & ConE           & 5.4           & 8.6           & 7.8           & \textbf{4.0}           & \textbf{3.6}           & 5.9           \\
                             & FLEX           & \textbf{5.6}  & \textbf{10.7} & \textbf{8.2}  & \textbf{4.0}  & \textbf{3.6}  & \textbf{6.5}  \\
      \midrule
      \multirow{3}{*}{NELL}  & BetaE          & 5.1           & 7.8           & 10.0          & 3.1           & 3.5           & 5.9           \\
                             & ConE           & 5.7           & 8.1           & 10.8          & 3.5           & 3.9           & 6.4           \\
                             & FLEX           & \textbf{5.8}  & \textbf{9.1}  & \textbf{10.9} & \textbf{3.6}  & \textbf{4.1}  & \textbf{6.7}  \\
      \bottomrule
    \end{tabular}
  }
  \vspace{-4mm}
\end{table*}



\xhdr{Impacts of Dimensionality}

\begin{wrapfigure}{r}{0.55\textwidth}
  \vspace{-16mm}
  \centering
  \includegraphics[width=0.45\textwidth]{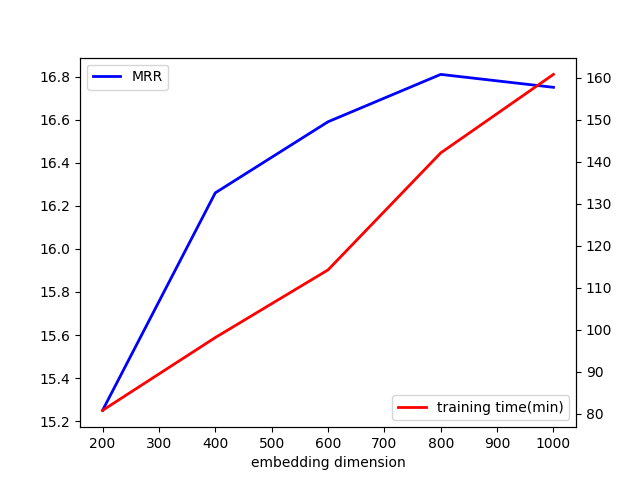}
  \caption{Impact of dimensionality on MRR and training time.}
  \label{fig:impact_of_dim}
\end{wrapfigure}
Our experiments indicate that the selection of embedding dimension has substantial influence on both effectiveness and efficiency of FLEX.
We train FLEX with embedding dimension $d\in\{200, 400, 600, 800, 1000\}$ and plot results based on the validation set, as shown in Figure~\ref{fig:impact_of_dim}.
With the increase of $d$, the training time rises, while the model performance (indicated by MRR) increases slowly during $d=200$ and $d=800$ but falls after $d=1000$.
Therefore, we decide 800 as the best setting.

\xhdr{Stable Performance}
We train and test FLEX five times with different random seeds and report the error bars in Appendix~\ref{sec:appendix:error_bars}. The small standard variances show that the performance of FLEX is stable.


\xhdr{Extensibility on Feature Part}
(1) Firstly, we show that the feature part can extend to be multiple.
We generate variant FLEX-2F which is FLEX with two feature parts and one logic part. Both feature parts depend on the same logic part, but they do not depend on each other.
In Table~\ref{table:variants}, Comparing to FLEX, FLEX-2F with two feature parts achieves up to 0.5\% MRR improvement on average, which demonstrates the extensibility of FLEX framework to add more feature parts.
We conclude that more parameters for the feature part result in more accurate reasoning answers, because it contributes to more precise description of entities.
(2) Secondly, we show that the feature part can be extended to be in infinite range ($L=\infty$).
We generate variant FLEX-$\infty$ which is FLEX with the real space as its feature space, infinite range. Specifically, $\tanh(\cdot)$ is removed in operators.
In Table~\ref{table:variants}, the result of FLEX-$\infty$ is quite close to FLEX's result on all tasks within 0.003 error.
Therefore, using the real space as the feature space is suitable, which shows the flexibility of the feature part in FLEX.

\xhdr{Flexibility on Logic Part}
We introduce min-based intersection and max-based union in Section~\ref{sec:model:extension} to show that our logic part is flexible to customize.
So, we generate variant FLEX-$\mathcal{I}_2$ which is FLEX using $\mathcal{I}_2$ as intersection operations.
In Table~\ref{table:variants}, both FLEX and FLEX-$\mathcal{I}_2$ achieve similar results on most tasks. Furthermore, FLEX-$\mathcal{I}_2$ gains the best result on $3i$ tasks.
The result demonstrates that different implementations of the logic part slightly affect the final performance.

\xhdr{The Improvement from Closure}
Modeling all closed logical operators allows the framework to train directly on queries with corresponding logical operations, eliminating the need for generalization ability from DNF or De Morgan (DM) formulation.
We design variant FLEX-$\mathcal{U}$, which is FLEX training on union queries. FLEX-$\mathcal{U}$ uses trainable union operator, instead of DNF or DM, to reason on queries containing union operation. In detail, we append 14968 newly generated \textbf{2u} and \textbf{up} queries to the training set. The validation set and test set remain the same. Note that the generated data doesn't contain any queries in validation set or test set.
From Table~\ref{table:variants}, we observe that FLEX-$\mathcal{U}$ achieves the best performance on \textbf{2u} and \textbf{up}, outperforming FLEX (which uses DNF) by 0.2\% and 0.6\% MRR, respectively.
Therefore, the closure of logical operators is helpful to improve the performance.

\xhdr{The Improvement from Logical Data}
We found that extra complex reasoning data is crucial to promote one-hop reasoning.
The variant FLEX-1p is FLEX trained on \textbf{1p} queries only.
Comparing it with standard FLEX in Table~\ref{table:variants}, FLEX-1p achieves the worst performance on \textbf{1p} task.
Besides, we further explore which task is more important for one-hop reasoning in Appendix~\ref{LP and MUL}.
The results show that logical queries is more important than multi-hop queries for the improvement of one-hop reasoning. Neural logical operators can capture the information hidden in logical data and contribute to the improvement of projection operator.

\xhdr{Ablation Study}
(1) If feature part is removed, all entities will be zeros, trivially, according to our definition in Section~\ref{sec:model:component} that the logic part of entity feature-logic embedding is zero.
(2) If the logic part is removed, all logic operators degenerate into simple neural networks. Then dyadic operators will be attention networks of the same structure, unable to guarantee to obey the rule of real logical operation.
We design variant FLEX-w/o-logic which removes logic part.
Overall, FLEX-w/o-logic is worse than FLEX across all metrics. Therefore, the logic part plays an important role in FLEX.

\xhdr{Complexity Analysis}
FLEX has fewer parameters and is faster than SOTA (BetaE, ConE), because the computation of logic parts is parameter-free. Please refer to Appendix~\ref{sec:appendix:complexity_analysis} for comparison about parameter and running time among different methods.

\xhdr{Case Study}
We conduct two case studies in Appendix~\ref{sec:appendix:case_study}.
\textbf{Case 1: The Visualization to Distinguish Answer Entities} shows that feature-logic framework is able to distinguish answer entities without the help of a strict geometric boundary like center-size framework.
\textbf{Case 2: The Visualization of Intermediate Answers in Reasoning} can help us to understand the reasoning process of FLEX and identify where errors may occur.
Because of space limitation, more research questions are presented in Appendix~\ref{sec:appendix:more_research_questions}.

\begin{table*}[!t]
  \vspace{-5mm}
  \caption{MRR comparison for all variants of FLEX answering queries with \& without negation on FB237. Each variant is based on basic FLEX (first row). The best results are in bold. \textbf{AVG} denotes average performance.
  }
  \label{table:variants}
  \centering
  \resizebox{\columnwidth}{!}{
    \begin{tabular}{l c ccccc ccccc cccc}
      \toprule
      \textbf{Model}       & \textbf{AVG}     & \textbf{1p}      & \textbf{2p}      & \textbf{3p}      & \textbf{2i}      & \textbf{3i}      & \textbf{ip}      & \textbf{pi}      & \textbf{2in}    & \textbf{3in}     & \textbf{inp} & \textbf{pin}    & \textbf{pni}    & \textbf{2u}      & \textbf{up}      \\
      \midrule
      FLEX                 & \underline{18.1} & \textbf{43.6}    & \underline{13.1} & \underline{11.1} & \underline{34.9} & 48.4             & 16.1             & \textbf{27.4}    & 5.6             & \textbf{10.7}    & 8.2          & \underline{4.0} & 3.6             & 15.4             & 11.1             \\
      \midrule
      FLEX-$\infty$        & \underline{18.1} & \underline{43.5} & \underline{13.1} & \underline{11.1} & \underline{34.9} & 48.6             & 16.4             & 26.9             & 5.6             & \underline{10.6} & 8.0          & 3.9             & 3.7             & \underline{15.5} & 11.2             \\
      FLEX-2F              & \textbf{18.3}    & \textbf{43.6}    & \textbf{13.3}    & \textbf{11.2}    & \textbf{35.0}    & \underline{48.8} & \textbf{17.4}    & \underline{27.3} & \textbf{5.8}    & 10.4             & \textbf{8.5} & \textbf{4.1}    & \textbf{4.0}    & \textbf{15.6}    & 11.5             \\
      \midrule
      FLEX-$\mathcal{I}_2$ & 18.0             & 43.3             & 12.8             & 10.9             & 34.8             & \textbf{49.4}    & 15.9             & 27.1             & 5.4             & 9.7              & 8.4          & \textbf{4.1}    & 3.6             & 14.9             & 10.9             \\
      \midrule
      FLEX-$\mathcal{U}$   & \underline{18.1} & 43.3             & 13.0             & \underline{11.1} & 34.5             & 47.7             & 16.6             & 27.0             & \textbf{5.8}    & 10.1             & \textbf{8.5} & \textbf{4.1}    & 3.8             & \textbf{15.6}    & \textbf{11.7}    \\
      FLEX-1p              & -                & 41.7             & -                & -                & -                & -                & -                & -                & -               & -                & -            & -               & -               & -                & -                \\
      \midrule
      FLEX-w/o-logic & 16.0          & 41.9          & 11.6          & 10.4          & 30.0          & 42.1          & 14.1          & 23.5          & 5.2          & 8.0           & 7.4          & 3.6          & 3.6          & 13.2          & 10.0          \\
      \bottomrule
    \end{tabular}
  }
  \vspace{-5mm}
\end{table*}


\section{Conclusion}\label{sec:conclusion}

In order to fill the gap that current embedding-based methods for KGR are coupled and not closed to the logic, we propose \textsl{F}eature-\textsl{L}ogic \textsl{E}mbedding framework, FLEX, by introducing vector logic for complex reasoning over knowledge graphs.
The theoretical analysis shows that it is a closed embedding-based framework for reasoning.
Additionally, the experimental results demonstrate that FLEX is a promising framework which has stable performance, extensibility, flexibility, and strong logical reasoning capability.

\begin{ack}
  This work was supported in part by the National Science Foundation of China (Grant No.61902034); Engineering Research Center of Information Networks, Ministry of Education of China

\end{ack}

{
\small
\bibliographystyle{unsrtnat}
\bibliography{paper_references/bibliography}

\begin{thebibliography}{24}
\providecommand{\natexlab}[1]{#1}
\providecommand{\url}[1]{\texttt{#1}}
\expandafter\ifx\csname urlstyle\endcsname\relax
  \providecommand{\doi}[1]{doi: #1}\else
  \providecommand{\doi}{doi: \begingroup \urlstyle{rm}\Url}\fi

\bibitem[Ren and Leskovec(2020)]{BetaE}
Hongyu Ren and Jure Leskovec.
\newblock Beta embeddings for multi-hop logical reasoning in knowledge graphs.
\newblock \emph{Neural Information Processing Systems (NeurIPS)}, 2020.

\bibitem[Choudhary et~al.(2021{\natexlab{a}})Choudhary, Rao, Katariya, Subbian,
  and Reddy]{PERM}
Nurendra Choudhary, Nikhil Rao, Sumeet Katariya, Karthik Subbian, and Chandan
  Reddy.
\newblock Probabilistic entity representation model for reasoning over
  knowledge graphs.
\newblock \emph{Advances in Neural Information Processing Systems}, 34,
  2021{\natexlab{a}}.

\bibitem[Ren et~al.(2020)Ren, Hu, and Leskovec]{Query2box}
H~Ren, W~Hu, and J~Leskovec.
\newblock Query2box: Reasoning over knowledge graphs in vector space using box
  embeddings.
\newblock In \emph{International Conference on Learning Representations
  (ICLR)}, 2020.

\bibitem[Zhang et~al.(2021)Zhang, Wang, Chen, Ji, and Wu]{ConE}
Zhanqiu Zhang, Jie Wang, Jiajun Chen, Shuiwang Ji, and Feng Wu.
\newblock Cone: Cone embeddings for multi-hop reasoning over knowledge graphs.
\newblock \emph{Advances in Neural Information Processing Systems}, 34, 2021.

\bibitem[Choudhary et~al.(2021{\natexlab{b}})Choudhary, Rao, Katariya, Subbian,
  and Reddy]{HypE}
Nurendra Choudhary, Nikhil Rao, Sumeet Katariya, Karthik Subbian, and Chandan~K
  Reddy.
\newblock Self-supervised hyperboloid representations from logical queries over
  knowledge graphs.
\newblock In \emph{Proceedings of the Web Conference 2021}, pages 1373--1384,
  2021{\natexlab{b}}.

\bibitem[Sun et~al.(2020)Sun, Arnold, Bedrax~Weiss, Pereira, and Cohen]{EmQL}
Haitian Sun, Andrew Arnold, Tania Bedrax~Weiss, Fernando Pereira, and William~W
  Cohen.
\newblock Faithful embeddings for knowledge base queries.
\newblock \emph{Advances in Neural Information Processing Systems}, 33, 2020.

\bibitem[Garg et~al.(2019)Garg, Ikbal, Srivastava, Vishwakarma, Karanam, and
  Subramaniam]{QuantumE}
Dinesh Garg, Shajith Ikbal, Santosh~K Srivastava, Harit Vishwakarma, Hima
  Karanam, and L~Venkata Subramaniam.
\newblock Quantum embedding of knowledge for reasoning.
\newblock \emph{Advances in Neural Information Processing Systems},
  32:\penalty0 5594--5604, 2019.

\bibitem[Srivastava et~al.(2020)Srivastava, Khandelwal, Madan, Garg, Karanam,
  and Subramaniam]{QuantumE+}
Santosh~Kumar Srivastava, Dinesh Khandelwal, Dhiraj Madan, Dinesh Garg, Hima
  Karanam, and L~Venkata Subramaniam.
\newblock Inductive quantum embedding.
\newblock \emph{Advances in Neural Information Processing Systems}, 33, 2020.

\bibitem[Wan and Du(2021)]{GaussianPath}
Guojia Wan and Bo~Du.
\newblock Gaussianpath: A bayesian multi-hop reasoning framework for knowledge
  graph reasoning.
\newblock In \emph{Proceedings of the AAAI Conference on Artificial
  Intelligence}, volume~35, pages 4393--4401, 2021.

\bibitem[Hamilton et~al.(2018)Hamilton, Bajaj, Zitnik, Jurafsky, and
  Leskovec]{GQE}
Will Hamilton, Payal Bajaj, Marinka Zitnik, Dan Jurafsky, and Jure Leskovec.
\newblock Embedding logical queries on knowledge graphs.
\newblock \emph{Advances in Neural Information Processing Systems},
  31:\penalty0 2026--2037, 2018.

\bibitem[Liu et~al.(2021)Liu, Du, Ji, Zhai, and Tong]{NewLook}
Lihui Liu, Boxin Du, Heng Ji, ChengXiang Zhai, and Hanghang Tong.
\newblock Neural-answering logical queries on knowledge graphs.
\newblock In \emph{Proceedings of the 27th ACM SIGKDD Conference on Knowledge
  Discovery \& Data Mining}, pages 1087--1097, 2021.

\bibitem[Arakelyan et~al.(2021)Arakelyan, Daza, Minervini, and Cochez]{CQD}
Erik Arakelyan, Daniel Daza, Pasquale Minervini, and Michael Cochez.
\newblock Complex query answering with neural link predictors.
\newblock In \emph{International Conference on Learning Representations}, 2021.
\newblock URL \url{https://openreview.net/forum?id=Mos9F9kDwkz}.

\bibitem[Bordes et~al.(2013)Bordes, Usunier, Garc{\'{\i}}a{-}Dur{\'{a}}n,
  Weston, and Yakhnenko]{TransE}
Antoine Bordes, Nicolas Usunier, Alberto Garc{\'{\i}}a{-}Dur{\'{a}}n, Jason
  Weston, and Oksana Yakhnenko.
\newblock Translating embeddings for modeling multi-relational data.
\newblock In \emph{NIPS 2013.}, 2013.

\bibitem[Sun et~al.(2019)Sun, Deng, Nie, and Tang]{RotatE}
Zhiqing Sun, Zhi-Hong Deng, Jian-Yun Nie, and Jian Tang.
\newblock Rotate: Knowledge graph embedding by relational rotation in complex
  space.
\newblock In \emph{International Conference on Learning Representations}, 2019.

\bibitem[Trouillon et~al.(2016)Trouillon, Welbl, Riedel, Gaussier, and
  Bouchard]{ComplEx}
Th{\'e}o Trouillon, Johannes Welbl, Sebastian Riedel, {\'E}ric Gaussier, and
  Guillaume Bouchard.
\newblock Complex embeddings for simple link prediction.
\newblock In \emph{International Conference on Machine Learning}, pages
  2071--2080. PMLR, 2016.

\bibitem[Chen et~al.(2022)Chen, Hu, and Sun]{FuzzQE}
X.~Chen, Ziniu Hu, and Yizhou Sun.
\newblock Fuzzy logic based logical query answering on knowledge graphs.
\newblock In \emph{AAAI}, 2022.

\bibitem[Luus et~al.(2021)Luus, Sen, Kapanipathi, Riegel, Makondo, Lebese, and
  Gray]{LogicE}
Francois Luus, Prithviraj Sen, Pavan Kapanipathi, Ryan Riegel, Ndivhuwo
  Makondo, Thabang Lebese, and Alexander Gray.
\newblock Logic embeddings for complex query answering.
\newblock \emph{arXiv preprint arXiv:2103.00418}, 2021.

\bibitem[Klement et~al.(2000)Klement, Mesiar, and Pap]{KlementTNormBook}
Erich{-}Peter Klement, Radko Mesiar, and Endre Pap.
\newblock \emph{Triangular Norms}, volume~8 of \emph{Trends in Logic}.
\newblock Springer, 2000.
\newblock ISBN 978-90-481-5507-1.
\newblock \doi{10.1007/978-94-015-9540-7}.
\newblock URL \url{https://doi.org/10.1007/978-94-015-9540-7}.

\bibitem[Mizraji(2008)]{vector_logic}
Eduardo Mizraji.
\newblock Vector logic: a natural algebraic representation of the fundamental
  logical gates.
\newblock \emph{Journal of Logic and Computation}, 18\penalty0 (1):\penalty0
  97--121, 2008.

\bibitem[Bollacker et~al.(2008)Bollacker, Evans, Paritosh, Sturge, and
  Taylor]{Freebase}
Kurt Bollacker, Colin Evans, Praveen Paritosh, Tim Sturge, and Jamie Taylor.
\newblock Freebase: a collaboratively created graph database for structuring
  human knowledge.
\newblock In \emph{Proceedings of the 2008 ACM SIGMOD international conference
  on Management of data}, pages 1247--1250, 2008.

\bibitem[Toutanova and Chen(2015)]{FB237}
Kristina Toutanova and Danqi Chen.
\newblock Observed versus latent features for knowledge base and text
  inference.
\newblock In \emph{3rd Workshop on Continuous Vector Space Models and Their
  Compositionality}, 2015.

\bibitem[Carlson et~al.(2010)Carlson, Betteridge, Kisiel, Settles, Hruschka,
  and Mitchell]{NELL}
Andrew Carlson, Justin Betteridge, Bryan Kisiel, Burr Settles, Estevam~R
  Hruschka, and Tom~M Mitchell.
\newblock Toward an architecture for never-ending language learning.
\newblock In \emph{Twenty-Fourth AAAI conference on artificial intelligence},
  2010.

\bibitem[Bala{\v{z}}evi{\'c} et~al.(2019)Bala{\v{z}}evi{\'c}, Allen, and
  Hospedales]{TuckER}
Ivana Bala{\v{z}}evi{\'c}, Carl Allen, and Timothy Hospedales.
\newblock Tucker: Tensor factorization for knowledge graph completion.
\newblock In \emph{Proceedings of the 2019 Conference on Empirical Methods in
  Natural Language Processing and the 9th International Joint Conference on
  Natural Language Processing (EMNLP-IJCNLP)}, pages 5185--5194, 2019.

\bibitem[Kingma and Ba(2015)]{Adam}
Diederick~P Kingma and Jimmy Ba.
\newblock Adam: A method for stochastic optimization.
\newblock In \emph{Proceedings of the International Conference on Learning
  Representations (ICLR)}, 2015.

\end{thebibliography}
}

\newpage
\appendix

\begin{center}
\begin{huge}
\textbf{Appendix}
\end{huge}
\end{center}

\section*{Broader Impact}
Multi-hop reasoning makes information stored in KGs more valuable.
With the help of multi-hop reasoning, we can digest more hidden and implicit information in KGs.
It will help in various application such as question answering, recommend systems and information retrieval.
It also shows a potential risk to expose unexpected personal information on public data.

\section{Proofs} \label{sec:appendix:theoretical_analysis}

\subsection{Proofs of Closure}

Our operators should be proven to be closed, because they involve neural network apart from vector logic.
We prove the closure of the operators in the following.

(Proposition~\ref{prop:Closure}) \textbf{Closure}: The projection operators, monadic and dyadic operators are closed under feature-logic embeddings of queries and entities.

\begin{proof}
  (Proposition~\ref{prop:Closure}) \textbf{Closure}:

  We divide the proof into three parts for the three types of operators.
  Each operator should be proved that the output should be feature-logic embeddings when given feature-logic embedding inputs.
  Formally, we should check (1) the output embedding consists of two parts, feature $\boldsymbol{\theta}_{f}'$ and logic $\boldsymbol{\theta}_{l}'$, and (2) the feature part is valid $\boldsymbol{\theta}_{f}'\in[-L,L]^d$ and the logic part is a valid logic embedding $\boldsymbol{\theta}_{l}'\in[0,1]^d$.
  The first is trivial. The logic part of the second is also trivial, because the logic part obeys vector logic and is closed.
  The feature part of the second is proved in the following.

  (1) For projection operator $\mathcal{P}$, it generates $\boldsymbol{\theta}_{f}'\in [-L, L]^d$ via activate function $g$.

  (2) For monadic operator $\mathcal{N}$, $\boldsymbol{\theta}_{f}' = L\tanh (\textbf{MLP}([\boldsymbol{\theta}_{f};\boldsymbol{\theta}_{l}]))\in [-L, L]^d$.

  (3) The feature part of dyadic operators is computed by
  \begin{equation*}
    \begin{aligned}
      \boldsymbol{\theta}_{f}' = & \sum_{i=1}^n \boldsymbol{\alpha}_i \boldsymbol{\theta}_{q_i,f}                                                                                                                     \\
      \boldsymbol{\alpha}_i =    & \frac{\exp (\textbf{MLP}([\boldsymbol{\theta}_{q_i,f};\boldsymbol{\theta}_{q_i,l}]))}{\sum_{j=1}^n \exp (\textbf{MLP}([\boldsymbol{\theta}_{q_j,f};\boldsymbol{\theta}_{q_j,l}]))}
    \end{aligned}
  \end{equation*}
  Notice that $\sum_{i=0}^{n} \boldsymbol{\alpha}_i = 1$ and $\boldsymbol{\theta}_{q_i,f} \in [-L,L]$. We have:
  \begin{equation*}
    |\boldsymbol{\theta}_{f}'| =|\sum_{i=1}^n \boldsymbol{\alpha}_i \boldsymbol{\theta}_{q_i,f}| \leq \sum_{i=1}^n \boldsymbol{\alpha}_i |\boldsymbol{\theta}_{q_i,f}| \leq \sum_{i=1}^n \boldsymbol{\alpha}_i L = L
  \end{equation*}
  Therefore $\boldsymbol{\theta}_{f}' \in [-L,L]^d$.

\end{proof}

\subsection{Proofs of Commutativity and Idempotence}
Our operators contain neural network apart from logic part. So the neural logical operators should be proven to also obey the rules of real set logical operations.
Here, we provide proofs for the following items:


(Proposition~\ref{prop:Commutativity}) \textbf{Commutativity}: Given feature-logic embedding $\mathcal{V}_{q_a}, \mathcal{V}_{q_b}$, we have $\mathcal{I}(\mathcal{V}_{q_a}, \mathcal{V}_{q_b}) = \mathcal{I}(\mathcal{V}_{q_b}, \mathcal{V}_{q_a})$ and $\mathcal{U}(\mathcal{V}_{q_a}, \mathcal{V}_{q_b}) = \mathcal{U}(\mathcal{V}_{q_b}, \mathcal{V}_{q_a})$, $\mathcal{I}_2(\mathcal{V}_{q_a}, \mathcal{V}_{q_b}) = \mathcal{I}_2(\mathcal{V}_{q_b}, \mathcal{V}_{q_a})$ and $\mathcal{U}_2(\mathcal{V}_{q_a}, \mathcal{V}_{q_b}) = \mathcal{U}_2(\mathcal{V}_{q_b}, \mathcal{V}_{q_a})$.

(Proposition~\ref{prop:Idempotence}) \textbf{Idempotence}: Given feature-logic embedding $\mathcal{V}_q$, we have $\mathcal{I}_2(\{\mathcal{V}_q,\mathcal{V}_q,\cdots,\mathcal{V}_q\}) = \mathcal{V}_q$ and $\mathcal{U}_2(\{\mathcal{V}_q,\mathcal{V}_q,\cdots,\mathcal{V}_q\}) = \mathcal{V}_q$.


\begin{proof}
  (Proposition~\ref{prop:Commutativity}) \textbf{Commutativity}:

  The definitions of the operator are listed as follows:

  Intersection Operator $\mathcal{I}$:

  \begin{equation*}
    \begin{aligned}
      \boldsymbol{\theta}_{f}' & =\sum_{i=1}^n \boldsymbol{\alpha}_i \boldsymbol{\theta}_{q_i,f} \\
      \boldsymbol{\theta}_{l}' & =\prod_{i=1}^n \boldsymbol{\theta}_{q_i,l}
    \end{aligned}
  \end{equation*}

  Union Operator $\mathcal{U}$:
  \begin{equation*}
    \begin{aligned}
      \boldsymbol{\theta}_{f}' = & \sum_{i=1}^n \boldsymbol{\alpha}_i \boldsymbol{\theta}_{q_i,f}                                 \\
      \boldsymbol{\theta}_{l}' = & \sum_{i=1}^n \boldsymbol{\theta}_{q_i,l}
      -\sum_{1\leqslant i < j\leqslant n} \boldsymbol{\theta}_{q_i,l} \boldsymbol{\theta}_{q_j,l}
      +\sum_{1\leqslant i < j < k\leqslant n} \boldsymbol{\theta}_{q_i,l} \boldsymbol{\theta}_{q_j,l} \boldsymbol{\theta}_{q_k,l} \\
                                 & +\dotsc +(-1)^{n-1} \prod_{i=1}^n \boldsymbol{\theta}_{q_i,l}
    \end{aligned}
  \end{equation*}
  where
  \begin{equation*}
    \boldsymbol{\alpha}_i = \frac{\exp (\textbf{MLP}([\boldsymbol{\theta}_{q_i,f};\boldsymbol{\theta}_{q_i,l}]))}{\sum_{j=1}^n \exp (\textbf{MLP}([\boldsymbol{\theta}_{q_j,f};\boldsymbol{\theta}_{q_j,l}]))}
  \end{equation*}

  $\mathcal{U}_2$ and $\mathcal{I}_2$ are the same with $\mathcal{U}$ and $\mathcal{I}$ except logic part.

  \begin{equation*}
    \begin{aligned}
      \mathcal{I}_2: & \boldsymbol{\theta}_{l}' =\min\{\boldsymbol{\theta}_{q_1,l},\boldsymbol{\theta}_{q_2,l},\cdots, \boldsymbol{\theta}_{q_n,l}\} \\
      \mathcal{U}_2: & \boldsymbol{\theta}_{l}' =\max\{\boldsymbol{\theta}_{q_1,l},\boldsymbol{\theta}_{q_2,l},\cdots, \boldsymbol{\theta}_{q_n,l}\} \\
    \end{aligned}
  \end{equation*}

  For the feature parts of all operations, the attention weights are invariant to permutations.

  For $\mathcal{U}_2$ and $\mathcal{I}_2$, the element-wise maximum function $\max(\cdot)$ and the element-wise minimum function $\min(\cdot)$ are also invariant to permutations.
  Thus, both $\mathcal{U}_2$ and $\mathcal{I}_2$ satisfy Commutativity.

  For $\mathcal{I}$, the intersection of logic part is naturally commutative, since multiplication is commutative.
  Therefore, $\mathcal{I}$ has Commutativity.

  For $\mathcal{U}$, the computation of logic parts is invariant to permutations, because both multiplication and addition comply to commutative law.
  So $\mathcal{U}$ is commutative as well.

\end{proof}

\begin{proof}
  (Proposition~\ref{prop:Idempotence}) \textbf{Idempotence}:

  Suppose that $\mathbf{V}_q=(\boldsymbol{\theta}_{f}, \boldsymbol{\theta}_{l})$.
  The attention weights $\alpha_i$ where $i=1,2,\cdots,n$ are the same, because each attention weight corresponds to the same query embedding $\mathcal{V}_q$ and the attention weights are invariant to permutations.
  Therefore, the feature part remains: $\boldsymbol{\theta}_{f}' =\sum_{i=1}^n \boldsymbol{\alpha}_i \boldsymbol{\theta}_{f} = (\sum_{i=1}^n \boldsymbol{\alpha}_i) \boldsymbol{\theta}_{f} = \boldsymbol{\theta}_{f}$.

  For $\mathcal{I}_2$, the logic part remains: $\boldsymbol{\theta}_{l}' =\min\{\boldsymbol{\theta}_{l},\boldsymbol{\theta}_{l},\cdots, \boldsymbol{\theta}_{l}\} = \boldsymbol{\theta}_{l}$.

  For $\mathcal{U}_2$, the logic part also remains: $\boldsymbol{\theta}_{l}' =\max\{\boldsymbol{\theta}_{l},\boldsymbol{\theta}_{l},\cdots, \boldsymbol{\theta}_{l}\} = \boldsymbol{\theta}_{l}$.

  To conclude, the operator $\mathcal{I}_2$ and $\mathcal{U}_2$ both have Idempotence.
\end{proof}






\section{Experiments Details} \label{sec:appendix:framework}

In this section, we show more details about our experiments, including Datasets Details~\ref{sec:appendix:datasets}, Implementation Details~\ref{sec:appendix:train} and Evaluation Details~\ref{sec:appendix:eval}.

\subsection{Datasets Details} \label{sec:appendix:datasets}
The datasets and query structured are created by \cite{BetaE}.
The statistics of the dataset are shown in Table \ref{table:datasets}.
The number of query for each dataset is shown in \ref{table:query}.

\begin{table*}
  \centering
  \caption{Dataset statistics}
  \label{table:datasets}
    \begin{tabular}{ccccccc}
      \toprule
      \textbf{Dataset} & \textbf{Entities} & \textbf{Relations} & \textbf{Training Edges} & \textbf{Validation Edges} & \textbf{Test Edges} & \textbf{Total Edges} \\
      \midrule
      FB15k            & 14,951            & 1,345              & 483,142                 & 50,000                    & 59,071              & 592,213              \\
      FB15k-237        & 14,505            & 237                & 272,115                 & 17,526                    & 20,438              & 310,079              \\
      NELL995          & 63,361            & 200                & 114,213                 & 14,324                    & 14,267              & 142,804              \\
      \bottomrule
    \end{tabular}
\end{table*}

\begin{table*}
  \caption{Number of query. EPFO represents query without negation. Neg represents query with negation. n1p represents query which is not 1p.}
  \label{table:query}
  \centering
  \begin{tabular}{ccccccc}
    \toprule
                     & \multicolumn{2}{c}{\textbf{Training}} & \multicolumn{2}{c}{\textbf{Validation}} & \multicolumn{2}{c}{\textbf{Test}}                                             \\
    \cmidrule(lr){2-3} \cmidrule(lr){4-5} \cmidrule(lr){6-7}
    \textbf{Dataset} & \textbf{EPFO}                         & \textbf{Neg}                            & \textbf{1p}                       & \textbf{n1p} & \textbf{1p} & \textbf{n1p} \\
    \midrule
    FB15k            & 273,710                               & 27,371                                  & 59,078                            & 8,000        & 66,990      & 8,000        \\
    FB237            & 149,689                               & 14,968                                  & 20,094                            & 5,000        & 22,804      & 5,000        \\
    NELL             & 107,982                               & 10,798                                  & 16,910                            & 4,000        & 17,021      & 4,000        \\
    \bottomrule
  \end{tabular}
\end{table*}

\subsection{Implementation Details} \label{sec:appendix:train}
We implement our model with PyTorch and use Adam\cite{Adam} as a gradient optimizer.
We use only one GTX1080 graphic card for experiments.
In order to find the best hyperparameters, we use grid search based on the performance on the validation datasets.
Please refer to Appendix~\ref{sec:appendix:train} for best hyperparameters.
The source code is available at~\repourl. We cite these projects and thank them for their great contributions!

\begin{enumerate}
  \item GQE (MIT License): \url{https://github.com/snap-stanford/KGReasoning}
  \item Query2box (MIT License): \url{https://github.com/snap-stanford/KGReasoning}
  \item BetaE (MIT License): \url{https://github.com/snap-stanford/KGReasoning}
  \item ConE (MIT License): \url{https://github.com/MIRALab-USTC/QE-ConE}
\end{enumerate}
The best hyperparameters of FLEX for each dataset are shown in Table  \ref{table:hyperparameters}.

\begin{table}
  \centering
  \caption{best hyperparameters on each dataset. $d$ is the embedding dimension, $b$ is the batch size, $k$ is the negative sampling size, $\gamma$ is the parameter in the loss function, $l$ is the learning rate.}
  \begin{tabular}{ccccc ccc}
    \toprule
    Dataset & $d$ & $b$ & $k$ & $\gamma$ & $l$               \\
    \midrule
    FB15k   & 800 & 512 & 128 & 30       & $1\times 10^{-4}$ \\
    FB237   & 800 & 512 & 128 & 30       & $1\times 10^{-4}$ \\
    NELL    & 800 & 512 & 128 & 30       & $1\times 10^{-4}$ \\
    \bottomrule
  \end{tabular}
  \label{table:hyperparameters}
\end{table}

\subsection{Evaluation Details}
\label{sec:appendix:eval}

For a test query $q$ and its non-trivial answer set $v \in \llbracket q \rrbracket_{test} - \llbracket q \rrbracket_{valid}$, we denote the rank of each answer $v_i$ as $rank(v_i)$.
The mean reciprocal rank (MRR) and Hits at K (Hits@K) are defined as follows:

\begin{equation}
  \begin{aligned}
    \text{MRR}(q)      & =\frac{1}{||v||}\sum_{v_i\in v}\frac{1}{rank(v_i)} \\
    \text{Hits@K}(q)   & =\frac{1}{||v||}\sum_{v_i\in v}f(rank(v_i))        \\
    \text{where } f(n) & =\begin{cases}
      1, & n \leq K \\
      0, & n > K
    \end{cases}
  \end{aligned}
\end{equation}

\section{Error Bars of Main Results} \label{sec:appendix:error_bars}

In order to evaluate how stable the performance of FLEX is, we run five times with random seeds $\{1, 10, 100, 1000, 10000\}$ and report the error bars of these results.
Table~\ref{table:main_results_error_bars_without_negation} shows the error bar of FLEX's MRR results on queries without negation (EPFO queries).
Table~\ref{table:main_results_error_bars_with_negation} shows the error bar of FLEX's MRR results on queries with negation.
Overall, the standard variances are small, which demonstrate that the performance of FLEX is stable.

\begin{table*}
  \caption{The mean values and standard variances of FLEX's MRR results for answering queries without negation ($\exists$, $\land$, $\lor$).}
  \label{table:main_results_error_bars_without_negation}
  \centering
    \begin{tabular}{c c c c c c c c c c c c}
      \toprule
      \textbf{Dataset}           & \textbf{1p}              & \textbf{2p}              & \textbf{3p}              & \textbf{2i}              & \textbf{3i}              & \textbf{pi}              & \textbf{ip}              & \textbf{2u}              & \textbf{up}              & \textbf{AVG}             \\
      \midrule
      \multirow{2}{*}{FB237-15k} &

      77.1                       & 37.4                     & 31.6                     & 66.4                     & 75.2                     & 54.2                     & 42.4                     & 52.9                     & 34.3                     & 52.4                                                \\

                                 & {\scriptsize $\pm$0.091} & {\scriptsize $\pm$0.021} & {\scriptsize $\pm$0.042} & {\scriptsize $\pm$0.011} & {\scriptsize $\pm$0.067} & {\scriptsize $\pm$0.120} & {\scriptsize $\pm$0.057} & {\scriptsize $\pm$0.122} & {\scriptsize $\pm$0.055} & {\scriptsize $\pm$0.024}

      \\

      \midrule
      \multirow{2}{*}{FB237}     &
      43.6                       & 13.1                     & 11.0                     & 34.9                     & 48.4                     & 27.4                     & 16.1                     & 15.4                     & 11.1                     & 24.6                                                \\
                                 & {\scriptsize $\pm$0.084} & {\scriptsize $\pm$0.097} & {\scriptsize $\pm$0.123} & {\scriptsize $\pm$0.152} & {\scriptsize $\pm$0.169} & {\scriptsize $\pm$0.096} & {\scriptsize $\pm$0.097} & {\scriptsize $\pm$0.184} & {\scriptsize $\pm$0.167} & {\scriptsize $\pm$0.025} \\

      \midrule
      \multirow{2}{*}{NELL}      &
      57.8                       & 16.8                     & 14.7                     & 40.5                     & 50.8                     & 27.3                     & 19.4                     & 15.6                     & 11.6                     & 28.2                                                \\

                                 & {\scriptsize $\pm$0.012} & {\scriptsize $\pm$0.071} & {\scriptsize $\pm$0.094} & {\scriptsize $\pm$0.114} & {\scriptsize $\pm$0.115} & {\scriptsize $\pm$0.130} & {\scriptsize $\pm$0.096} & {\scriptsize $\pm$0.026} & {\scriptsize $\pm$0.118} & {\scriptsize $\pm$0.032} \\
      \bottomrule
    \end{tabular}
\end{table*}

\begin{table*}[t]
  \caption{The mean values and standard variances of FLEX's MRR results for answering queries with negation .}
  \label{table:main_results_error_bars_with_negation}
  \centering
  \begin{tabular}{ c c  c c c c  c c c c }
    \toprule
    \textbf{Dataset} & \textbf{2in}             & \textbf{3in}             & \textbf{inp}             & \textbf{pin}             & \textbf{pni}             & \textbf{AVG}             \\
    \midrule
    \multirow{2}{*}{FB15k}
                     & 18.0                     & 19.3                     & 14.2                     & 10.1                     & 15.2                     & 15.4                     \\
                     & {\scriptsize $\pm$0.073} & {\scriptsize $\pm$0.089} & {\scriptsize $\pm$0.059} & {\scriptsize $\pm$0.045} & {\scriptsize $\pm$0.087} & {\scriptsize $\pm$0.051} \\
    \midrule
    \multirow{2}{*}{FB237}
                     & 5.6                      & 10.7                     & 8.3                      & 3.9                      & 3.6                      & 6.1                      \\
                     & {\scriptsize $\pm$0.032} & {\scriptsize $\pm$0.119} & {\scriptsize $\pm$0.055} & {\scriptsize $\pm$0.065} & {\scriptsize $\pm$0.052} & {\scriptsize $\pm$0.037} \\
    \midrule
    \multirow{2}{*}{NELL}
                     & 5.8                      & 9.1                      & 10.9                     & 3.6                      & 4.1                      & 6.7                      \\
                     & {\scriptsize $\pm$0.015} & {\scriptsize $\pm$0.057} & {\scriptsize $\pm$0.014} & {\scriptsize $\pm$0.013} & {\scriptsize $\pm$0.092} & {\scriptsize $\pm$0.021} \\
    \bottomrule
  \end{tabular}
\end{table*}


\section{Complexity Analysis} \label{sec:appendix:complexity_analysis}

Because the computation of logic parts is parameter-free, FLEX has less parameters and is faster than SOTA (BetaE, ConE).
The parameter comparison is shown in Table~\ref{table:parameter_comparison} and the running time comparison is presented in Table~\ref{table:running_time_comparison}. From the tables, we observe that FLEX has the least parameters and running time. These results demonstrate the efficiency of FLEX framework.

\begin{table*}[!t]
  \caption{Parameters of BetaE, ConE and FLEX, FLEX-2F, FLEX-3F along different hidden dimensions. The one having least parameters is in bold.}
  \label{table:parameter_comparison}
  \centering
  \begin{tabular}{c c c c c c}
    \toprule
    \textbf{Dimension} & \textbf{BetaE} & \textbf{ConE} & \textbf{FLEX}   & \textbf{FLEX-2F} & \textbf{FLEX-3F} \\
    \midrule
    200                & 10.30M         & 7.18M         & \textbf{7.05M}  & 10.81M           & 14.57M           \\
    400                & 18.52M         & 12.27M        & \textbf{11.79M} & 19.54M           & 27.29M           \\
    600                & 27.22M         & 17.84M        & \textbf{16.76M} & 28.75M           & 40.74M           \\
    800                & 36.39M         & 23.89M        & \textbf{21.97M} & 38.43M           & 54.89M           \\
    1000               & 46.05M         & 30.42M        & \textbf{27.42M} & 48.60M           & 69.78M           \\
    \bottomrule
  \end{tabular}
\end{table*}

\begin{table*}[!t]
  \caption{Running time of BetaE, ConE and FLEX, FLEX-2F, FLEX-3F along different hidden dimensions. They are trained on one 1080Ti card, batch size 128, training steps 300k. The one having least days is in bold.}
  \label{table:running_time_comparison}
  \centering
  \begin{tabular}{c c c c c c}
    \toprule
    \textbf{Dimension} & \textbf{BetaE} & \textbf{ConE} & \textbf{FLEX}     & \textbf{FLEX-2F} & \textbf{FLEX-3F} \\
    \midrule
    200                & 3.5 days       & 1.9 days      & \textbf{1.2 days} & 2.2 days         & 3.3 days         \\
    400                & 4.1 days       & 2.1 days      & \textbf{1.3 days} & 2.5 days         & 3.8 days         \\
    600                & 4.4 days       & 2.3 days      & \textbf{1.5 days} & 2.7 days         & 4.0 days         \\
    800                & 4.6 days       & 2.4 days      & \textbf{1.7 days} & 2.9 days         & 4.3 days         \\
    1000               & 4.9 days       & 2.6 days      & \textbf{1.8 days} & 3.1 days         & 4.5 days         \\
    \bottomrule
  \end{tabular}
\end{table*}

\section{Case Study} \label{sec:appendix:case_study}

\begin{figure*}
  \begin{subfigure}[b]{0.32\textwidth}
    \includegraphics[width=\textwidth]{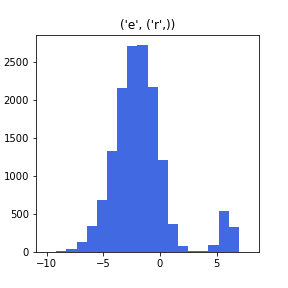}
    \caption{1p task}
  \end{subfigure}
  \hfill
  \begin{subfigure}[b]{0.32\textwidth}
    \includegraphics[width=\textwidth]{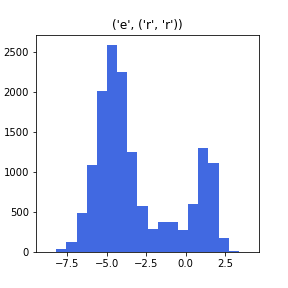}
    \caption{2p task}
  \end{subfigure}
  \hfill
  \begin{subfigure}[b]{0.32\textwidth}
    \includegraphics[width=\textwidth]{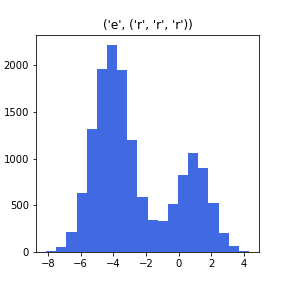}
    \caption{3p task}
  \end{subfigure}
  \caption{The count of entities grouped by scores on different tasks. Horizontal axis denotes the scores, while vertical axis denotes the number of entity in the score interval.}
  \label{fig:case study}
\end{figure*}

\subsection{Case 1: The Visualization to Distinguish Answer Entities}

To investigate whether the feature-logic framework can distinguish answer entities from other entities, we test the framework with three tasks \textbf{1p}, \textbf{2p},\textbf{3p} on FB15k.
We randomly select three samples: \textbf{1p} ('/m/0jbk9', ('-/location/ hud\_foreclosure\_area/ estimated\_number\_of\_mortgages./ measurement\_unit/ dated\_integer/ source',)), \textbf{2p} ('/m/02x\_h0', ('+/ people/ person/ spouse\_s./ people/ marriage/ type\_of\_union', '+/ people/ marriage\_union\_type/ unions\_of\_this\_type./ people/ marriage/ spouse')), \textbf{3p} ('/m/02h40lc', ('-/ location/ country/ official\_language', '+/ dataworld/ gardening\_hint/ split\_to', '-/ people/ person/ nationality')).
Figure~\ref{fig:case study} shows the entity count along different ranges of scores.
Horizontal axis denotes the scores, while vertical axis denotes the number of entity in the score interval.
Higher score means higher confidence for the entity as answer.

We can observe two obvious peaks in each figure.
The left peak represents the number of entities which aren't answers and the right peak represents the number of entities which are considered answers.
It can be concluded that feature-logic framework is able to distinguish answer entities without the help of a strict geometric boundary like center-size framework.

Besides, with the increase in the complexity of the task (from \textbf{1p} to \textbf{3p}), the right peak becomes higher and higher, which means the number of answers inferred increase. Too many answer entities inferred will confuse the framework and lead to performance decrease.

\subsection{Case 2: The Visualization of Intermediate Answers in Reasoning}

The intermediate variables in FLEX are interpretable.
Each intermediate node is represented as feature-logic embedding.
The feature-logic embedding is used to rank against all candidate entities to get the answers of the previous reasoning step.
We visualize the top-3 entities with the highest scores.

\begin{figure*}
  \centering
  \includegraphics[width=\textwidth]{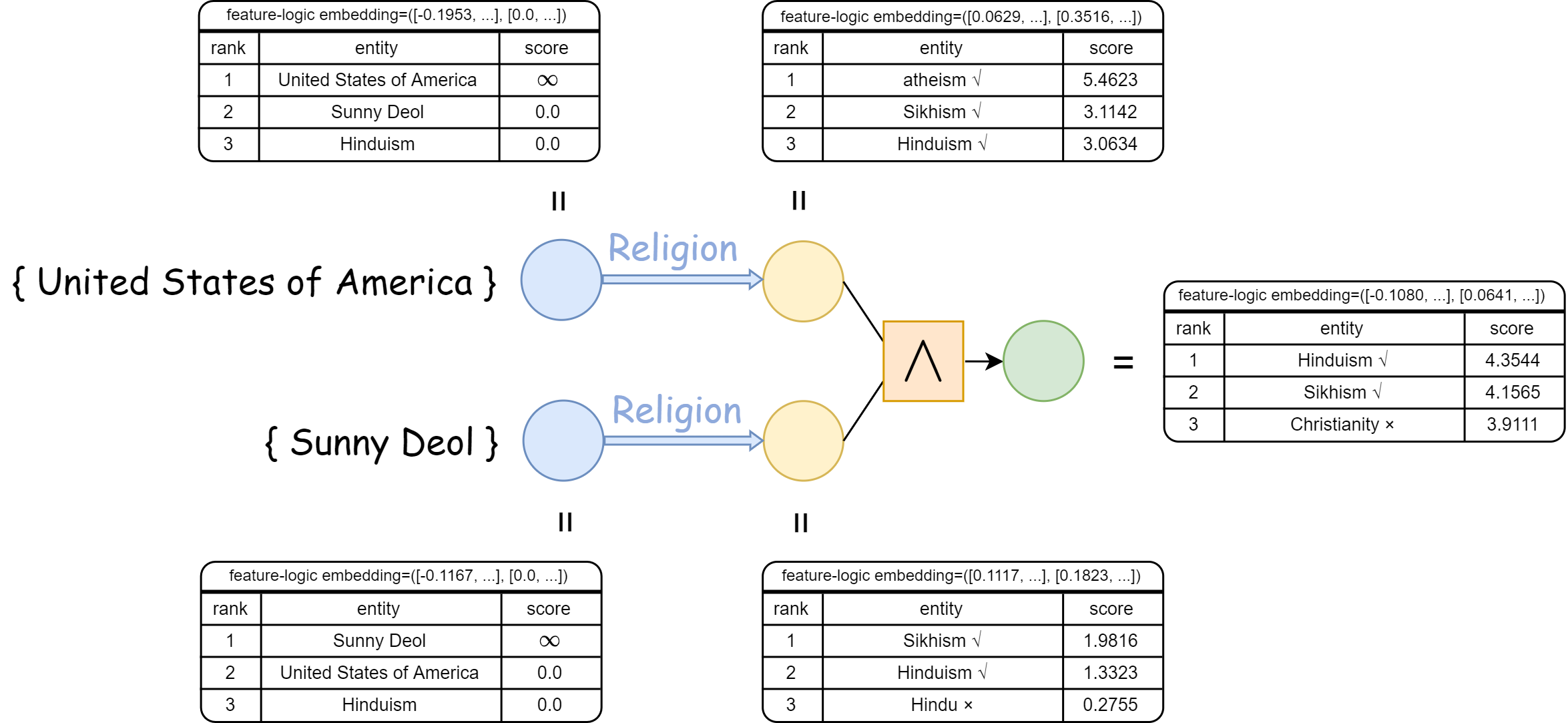}
  \caption{The visualization of the intermediate answers in reasoning, given by \textbf{2i} query $q=V_?: (\text{United States of America}, \text{Religion}, V_?) \land (\text{Sunny Deol}, \text{Religion}, V_?)$.}
  \label{fig:case study 2}
\end{figure*}

We randomly choose a \textbf{2i} query from the FB237 dataset.
Figure~\ref{fig:case study 2} shows the visualization of the intermediate answers in the reasoning process using the FLEX framework.
At the anchor step, the maximum score is given to anchor entity itself, since the distance between the anchor and itself is 0.
In the last step, FLEX correctly identifies the final answers of 'Hinduism' and 'Sikhism,' and ignores the incorrect answer of 'Hindu' that was generated in previous steps.
However, FLEX also assigns a relatively high score to another entity 'Christianity'.
'Christianity' could serve as a good suggestion for the user to check whether it was answer.

Overall, since the ground truth is 'Hinduism' and 'Sikhism,' the FLEX framework performs well in this case.
The intermediate answers generated by the framework are reasonable and can be used to infer the final answer.
The visualization of the intermediate answers can help us to understand the reasoning process of FLEX and identify where errors may occur.

\section{More Research Questions} \label{sec:appendix:more_research_questions}

In this section, we designed more variants to further explore how powerful the FLEX framework is.

Mainly we concern the research questions: (RQ1) What if the logic operations are directly applied on the feature parts? (RQ2) What if we make it 3 or more feature parts? What is the cost of adding a more features spaces in the framework, in terms of training time/memory requirements etc?

Below, the variant \textbf{FLEX-logic-over-feature} is designed so that logic operations are applied directly on the feature part without any neural networks. The variant \textbf{FLEX-3F} has three feature parts, and \textbf{FLEX-2F} has two.
The results are shown in Table~\ref{table:more_variants}.

\begin{table*}[!t]
  \caption{MRR comparison for more variants of FLEX answering queries with \& without negation on FB237. Each variant is based on basic FLEX (first row). \textbf{AVG} denotes average performance.}
  \label{table:more_variants}
  \centering
  \resizebox{\textwidth}{!}{
    \begin{tabular}{l c ccccc ccccc cccc}
      \toprule
      \textbf{Model}          & \textbf{AVG} & \textbf{1p} & \textbf{2p} & \textbf{3p} & \textbf{2i} & \textbf{3i} & \textbf{ip} & \textbf{pi} & \textbf{2in} & \textbf{3in} & \textbf{inp} & \textbf{pin} & \textbf{pni} & \textbf{2u} & \textbf{up} \\
      \midrule
      FLEX                    & 18.1         & 43.6        & 13.1        & 11.1        & 34.9        & 48.4        & 16.1        & 27.4        & 5.6          & 10.7         & 8.2          & 4.0          & 3.6          & 15.4        & 11.1        \\
      FLEX-logic-over-feature & 13.2         & 40.1        & 10.6        & 9.6         & 24.9        & 36.8        & 11.6        & 14.9        & 5.3          & 6.2          & 6.7          & 3.5          & 3.8          & 11.2        & 8.8         \\
      FLEX-2F                 & 18.3         & 43.6        & 13.3        & 11.2        & 35.0        & 48.8        & 17.4        & 27.3        & 5.8          & 10.4         & 8.5          & 4.1          & 4.0          & 15.6        & 11.5        \\
      FLEX-3F                 & 18.4         & 43.6        & 13.3        & 11.2        & 35.2        & 49.0        & 17.5        & 27.4        & 5.9          & 10.5         & 8.5          & 4.2          & 4.1          & 15.7        & 11.1        \\
      \bottomrule
    \end{tabular}
  }
\end{table*}

\paragraph{RQ1}
Compared with basic FLEX, the variant \textbf{FLEX-logic-over-feature} has significant performance decrease. Therefore, we conclude that Neural logical operators do better in handling noise in the KG, and FLEX framework is more suitable to fuse the logical information into neural networks.

\paragraph{RQ2}
Performance improvement is limited for three or more feature parts compared with two feature parts. Intuitively, too many feature parts may carry in more parameters (making it hard to train) and more noises (leading to ignoring the logic). We suggest that the neural networks in the operators should be redesigned and promoted to well handle these features.

Below we analyze the computation cost of adding a more features spaces in the framework. Let $n$ denotes number of entities, $m$ denotes number of relations, $d$ denotes embedding dimension, $h$ denotes hidden dimension. \textbf{If we add a feature part, there are about $(n + m)d + 2hd + 3d^2$ extra parameters introduced, and the training time increases linearly.}

(1)  Each entity embedding and relation embedding needs extra $d$ dimension parameters, which leads to extra $(n + m) \times d$ parameters.

(2)  The projection operation contains an 2 layer feedforward network of size $(x + 1)d \times h$ and $h \times (x + 1)d$, where $x$ is the number of feature parts, 1 is one logic part, $h$ is hidden dimension. Adding  a feature part requires $2hd$ extra parameters.

(3)  For each feature part, intersection operation needs an additional 2 layer feedforward network, which requires first layer $2d \times d$ and second layer $d \times d$ summing up to $3d^2$ extra parameters.

By the way, FLEX-2F and FLEX-3F's results of parameters and running time along different hidden dimensions are shown in Table~\ref{table:parameter_comparison} and Table~\ref{table:running_time_comparison}.

\section{Link Prediction and Multi-Hop Reasoning} \label{LP and MUL}

Reasoning on KG is a fundamental issue in the artificial intelligence which can be classified into two categories:
(1) Link prediction, which uses existing facts stored in KG to infer new one,
and (2) Multi-hop reasoning, which answers a first-order logic (FOL) query using logical and relational operators.
Multi-hop reasoning is more difficult than link prediction, as it involves multiple entities, relationships.
However, on the one hand, current SOTA models for link prediction cannot handle multi-hop reasoning queries.
On the other hand, the SOTA methods for multi-hop reasoning cannot outperform SOTA approaches for link prediction.
Our framework not only can handle multi-hop queries, but also achieves comparable results to SOTA models for link prediction.

We compare FLEX with ConE (the multi-hop SOTA) as well as TuckER (the link prediction SOTA).
For ConE, we reproduce and obtain its MRR and Hits@k results on 1p task.
For TuckER, We include its tail prediction metrics in the comparison.
Note that the 1p task in multi-hop reasoning is the same as tail prediction task in link prediction.

From Table~\ref{table:LP multi}, it can be observed that:

(1) \textbf{FLEX} and its variants outperform current SOTA multi-hop reasoning model (ConE) in link prediction task and achieves comparable result with current SOTA link prediction model (TuckER).

(2) Multi-hop training contributes to the framework's one-hop performance. \textbf{FLEX} trained with complex multi-hop data outperforms the model \textbf{FLEX-1p} trained only with 1p data.

\begin{table}
  \centering
  \caption{The metrics of one-hop reasoning (Link Prediction) on FB237, ConE~\cite{ConE} and TuckER~\cite{TuckER} are reproduced. The best is in bold.}
  \label{table:LP multi}
  \begin{tabular}{cccc}
    \toprule
    \multirow{2}{*}{Metric} & \multicolumn{3}{c}{Model}                        \\
                            & FLEX                      & ConE & TuckER        \\
    \midrule
    MRR                     & 43.6                      & 41.8 & \textbf{45.1} \\
    hits@1                  & 33.1                      & 31.5 & \textbf{35.4} \\
    hits@3                  & 48.4                      & 46.5 & \textbf{49.7} \\
    hits@10                 & \textbf{64.9}             & 62.6 & 64.0          \\
    \bottomrule
  \end{tabular}
\end{table}

\paragraph{Why is FLEX better in 1p?}
As is discussed before, the training on logical queries contributes to the performance of projection operator. Below we conduct experiments to explore the reason.
To answer this question, we further train FLEX on the same dataset FB237 of 4 different settings: \textbf{1p}, \textbf{1p.2p.3p}, \textbf{1p.2p.3p.2i.3i}, \textbf{all} respectively. Note that the result of training on \textbf{all} is the result of FLEX presented before in the main body of the paper. The results on 1p task are shown in Table~\ref{table:1p}:

\begin{table*}
  \centering
  \caption{Experiment results on 1p task (which equals to Link Prediction) for FLEX trained on FB237 using queries: 1p (\textbf{FLEX-1p}), 1p.2p.3p (\textbf{FLEX-1p.2p.3p}), 1p.2p.3p.2i.3i (\textbf{FLEX-1p.2p.3p.2i.3i}) and all (\textbf{FLEX}) respectively.. The best is in bold.}
  \label{table:1p}
  \begin{tabular}{ccccc}
    \toprule
    Metric  & \textbf{FLEX-1p} & \textbf{FLEX-1p.2p.3p} & \textbf{FLEX-1p.2p.3p.2i.3i} & \textbf{FLEX} \\
    \midrule
    MRR     & 41.7             & 41.0                   & 42.8                         & \textbf{43.6} \\
    hits@1  & 31.9             & 31.0                   & 32.5                         & \textbf{33.1} \\
    hits@3  & 46.1             & 45.3                   & 47.4                         & \textbf{48.4} \\
    hits@10 & 61.6             & 61.1                   & 63.8                         & \textbf{64.9} \\
    \bottomrule
  \end{tabular}
\end{table*}

From Table~\ref{table:1p}, the MRR decreases from \textbf{1p}(41.7) to \textbf{1p.2p.3p}(41.0) then increases as more and more logical data involved, from \textbf{1p.2p.3p}(41.0), \textbf{1p.2p.3p.2i.3i}(42.8) to \textbf{all}(43.6). Extra 2p.3p data brings in more data instead of introducing logical operators, then the performance decreases. So extra data doesn’t always contribute to performance. But when using 2i.3i data and including intersection operator, the 1p performance of FLEX increases significantly. Therefore, \textbf{neural logical operators contribute to the improvement of projection operator.} The result inspires us that FLEX framework could well-organize the logical operators and capture the information hidden in logical data.

\section{Connection to Other Works}

\subsection{Connection to ConE} \label{FLEX and ConE}

ConE~\cite{ConE} models the query set as sector cone.
It shows impressive performance in reasoning over knowledge graph.
The difference between ConE and our FLEX can be listed as follows:

(1) ConE's union is not true union because the union of sector cones is cone instead of sector cone. FLEX's union is true union in vector logic. In fact, including Q2B~\cite{Query2box}, if center-size framework accurately models any EPFO logical queries, the dimensionality of the logical query embeddings needs to be $\mathcal{O}(|\mathcal{V}|)$, where $|\mathcal{V}|$ is the count of all entities. And this is why we would like to propose a method to directly model union in vector logic. The good result of experiment (\textbf{FLEX-$\mathcal{U}$}) shows the necessary of union operator.

(2) The axis and aperture in ConE depend on each other.
In FLEX, the feature and logic are more loosely decoupled.
Theoretically, FLEX allows integrating multi-modal information including entity name, image, audio, etc. Please refer to \textbf{FLEX-2F} to add another feature part.
However, the experiments remain a future work, because we do not have a multi-modal dataset.

\subsection{Connection to CQD} \label{FLEX and CQD}

CQD~\cite{CQD} firstly trains a neural link predictor to predict truth value of atoms in the query. It then uses t-norm to organise the truth value of different atoms. Finally it solves a optimization problem to find the answer entity set.
The similarity and difference between CQD and FLEX are listed as follows.

(1) Similarity: CQD uses t-norm while FLEX uses vector logic to model logical operators in the query.
T-norm and vector logic both originate from fuzzy logic, so we share similar definitions of \textbf{AND} and \textbf{OR} operations with CQD.
The product t-norm and minimum t-norm correspond to our intersection operator $\mathcal{I}$ and $\mathcal{I}_2$ respectively.
And the probabilistic sum and maximum t-conorm correspond to our union operator $\mathcal{U}$ and $\mathcal{U}_2$ respectively.

(2) Difference: The difference between CQD and FLEX is training time and supported FOL operators.
CQD utilizes t-norm to promote neural link predictor to complex logical reasoning over knowledge graph.
Therefore, CQD has the advantage to train on \textbf{1p} tasks, while testing in all tasks (without negation).
This advantage makes the training of CQD much faster than our FLEX.
However, in the paper of CQD, neither the definition of negation of t-norm or the result of queries with negation is shown.
Our FLEX can not only represent negation (\ie~NOT operation in vector logic), but also represent all dyadic operations including exclusive or (XOR), implication (IMPL) and so on.
Actually, FLEX can truly directly handle all FOL operations.

\section{Future Work}

In the future, we would like to explore the following research directions: (1) We plan to optimize the training pipeline to reduce training time; (2) We would like to promote FLEX to temporal knowledge graph in the future; (3) We aim to investigate application of the quantum logic as the computation of logic parts.

\end{document}